\documentclass{article}

 \usepackage[main, final]{neurips_2025}

\usepackage[utf8]{inputenc} 
\usepackage[T1]{fontenc}    
\usepackage{hyperref}       
\usepackage{url}            
\usepackage{booktabs}       
\usepackage{amsfonts}       
\usepackage{nicefrac}       
\usepackage{float}

\usepackage[table]{xcolor}
\usepackage{xcolor}         

\usepackage{times}
\usepackage{latexsym}
\usepackage{array}
\usepackage{multicol}
\usepackage{tablefootnote}
\usepackage{threeparttable}
\usepackage{pifont}
\usepackage{lipsum}
\setlipsum{%
  par-before = \begingroup\color{gray},
  par-after = \endgroup
}

\usepackage{graphicx}
\usepackage{amsmath}
\usepackage{mdframed}
\usepackage{subcaption}

\usepackage{xspace}
\usepackage{kotex} 

\usepackage{listings}  

\usepackage{adjustbox}
\usepackage[breakable]{tcolorbox}

\usepackage{multirow}
\usepackage{svg}
\usepackage{titlesec}

\usepackage{bbm}
\usepackage{subcaption}
\usepackage{bm}

\usepackage{todonotes}      
\usepackage{natbib}

\usepackage{tabularx}
\usepackage{makecell}

\definecolor{customblue}{HTML}{6BE2F6}

\newcommand{\hcx}{HyperCLOVA X\xspace}
\newcommand{\hcxtfull}{HyperCLOVA X THINK\xspace}
\newcommand{\hcxt}{THINK\xspace}

\newcommand{\shortsection}[1]{\noindent\textbf{#1.}}

\title{\hcxtfull}

\author{NAVER Cloud \\ HyperCLOVA X Team}

\begin{document}

\maketitle
\begin{abstract}

We introduce \hcxtfull, the first reasoning-focused large language model in the \hcx family, pre-trained on roughly $6$ trillion high-quality Korean, and English tokens, augmented with targeted synthetic Korean data. 
 It was implemented as a compute-memory-balanced Peri-LN Transformer scaled with $\mu$P, pre-trained through a three-stage curriculum that expands the context window to $128$K tokens,
and post-trained via supervised fine-tuning with Reinforcement Learning from Verifiable Rewards supports both detailed rationale and concise-answer modes. 
It delivers competitive performance against similarly sized models on Korea-focused benchmarks such as KMMLU, CSAT, KoBALT-700, HAERAE-1.0, and KoBigBench, while preserving robust bilingual consistency and translation quality. In addition, a vision-augmented variant matches or exceeds GPT-4.1 on the KCSAT STEM benchmark, all of which are achieved with substantially lower training compute than existing models of similar sizes.  
These capabilities position \hcxtfull as a robust foundation for Korean AI innovation and a valuable resource for the global research community.
Lastly, we present a pruning and distillation technique that will soon be applied to \hcxtfull for an open-source and business-friendly foundation model.

\end{abstract}
\section{Introduction}
Recent advancements of large language models (LLMs) have drawn increased attention to their reasoning abilities, going beyond simple memorization of factual knowledge to deriving logical conclusions. Models like GPT-o1~\citep{openai2024openaio1card}, R1~\citep{deepseekai2025deepseekr1incentivizingreasoningcapability}, and QwQ~\citep{qwq32b} exemplify such effort, demonstrating that the ability to perform logical inferences and multi-step problem solving can significantly broaden the scope of AI applications.

At the same time, the notion of sovereign AI is being established as an important goal. As LLMs continue to be deployed in various regions around the globe, there is a growing need for linguistic fluency and cultural sensitivity tailored toward a given region, as well as data governance that aligns with regional values and regulations. In this regard, our immediate focus is Korea.

To meet the imperatives of both advanced reasoning and sovereign AI---for Korea, in particular---we present \hcxtfull(\textit{henceforth} \hcxt).
It is the first reasoning-focused LLM in the \hcx family\citep{yoo2024hyperclovaxtechnicalreport}, trained via a strategic preparation of training data and use of the latest pre- and post-training techniques.

In particular, we curated a corpus of roughly six trillion tokens that balances high-quality Korean and English text with targeted synthetic Korean data. This mixture improves linguistic breadth while safeguarding cultural and domain relevance. 
The model architecture follows a compute-memory-balanced Peri-LN Transformer scaled with the $\mu$P framework, allowing consistent hyperparameter transfer from small to large scales without extensive grid search. 

During pre-training, A three-stage curriculum gradually increases the context window, culminating in 128k tokens, which enables THINK to process long documents and perform multi-step reasoning within a single pass. Then, for post-training, we combine supervised fine-tuning on carefully designed reasoning tasks with Reinforcement Learning from Verifiable Rewards. This alignment strategy encourages the model to generate explicit chains of thought when requested and concise answers when brevity is preferred. Safety alignment follows NAVER AI Ethics guidelines through filtered data, red-teaming, refusal sampling, and policy tuning.

We evaluate THINK on Korea-focused benchmarks such as KMMLU, CSAT, KoBALT-700, HAERAE-1.0, and KoBigBench. The model achieves competitive accuracy among similarly sized models while requiring substantially lower training compute. 
A vision-augmented variant that integrates vision encoders to extend the same reasoning framework to image-text tasks, matches or surpasses GPT-4.1 on the KCSAT STEM benchmark.

To ensure that academic and industry partners can benefit from the model, we introduce a pruning-and-distillation recipe that reduces parameter count while preserving accuracy. This technique will soon be applied to THINK itself to produce a model suitable for limited resource settings. We plan to open-source release this model under a business-friendly license. 

Our contributions are threefold. First, we demonstrate that a regionally tailored corpus combined with modern scaling laws yields a bilingual model with strong reasoning capability. Second, we provide an efficient training and alignment recipe that lowers the barrier to entry for sovereign AI development. Third, we share a practical pruning-distillation pipeline and commit to apply it for an open-source version of THINK---fostering further research and commercial deployment, even under more resource-constrained settings.


\section{Pre‑Training}
\begin{figure}[tp!]
    \centering
    \includegraphics[width=\textwidth]{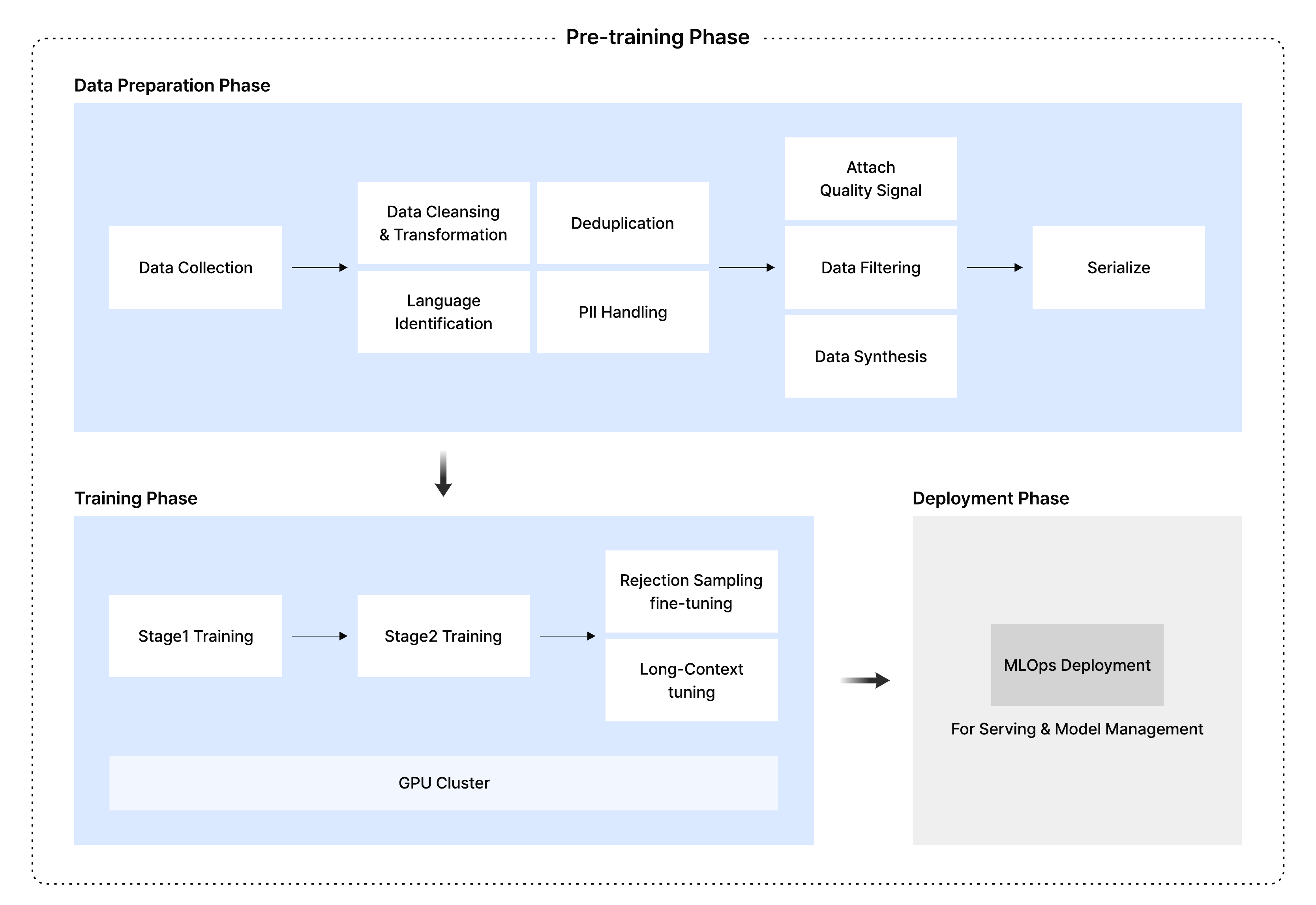}
\caption{Pre-training pipeline of \hcxtfull. (1) Data-Preparation Phase: A scalable pipeline collects raw corpora, carries out cleansing, language identification, deduplication, and masking; attaches quantitative quality signals, applies filtering, synthesizes targeted data, and serializes the resulting shards (2) Training Phase: A dedicated three-stage curriculum, with each stage optimized for its specific objective, progressively builds and refines the model’s capabilities. 
}
    \label{fig:overview_pre_training}
\end{figure}

This section outlines the pre-training methodology behind \hcxt: a scalable, Korean-centric data pipeline enriched with targeted synthetic corpora (Section \ref{pre-training:data}); a compute–memory‑efficient yet stability‑oriented Transformer, instantiated with scale‑invariant parameterization principles (Section \ref{pre-training:architecture}); and a three-stage curriculum that sequentially builds foundational linguistic knowledge, refines competence with higher-fidelity data, and expands contextual capacity to support long-form reasoning  (Section \ref{pre-training:stage}).
See Figure \ref{fig:overview_pre_training} for an overview of the pre-training process.

\subsection{Data Preparation} \label{pre-training:data}  
We begin with the end-to-end data pipeline---collection, cleaning, and quality filtering---paying special attention to techniques tailored for our large-scale Korean corpus. We then describe a synthetic-data generation strategy that enriches under-represented domains while preserving linguistic fidelity. 

\shortsection{Data Pipeline}
The data pipeline for \hcxt is designed around three guiding principles: scalability, reusability, and quick refresh, so that new corpora can be incorporated with minimal latency while maintaining strict quality guarantees. Following \citet{redpajama}, the pipeline separates schema standardization from quality assessment and filtering. During standardization, raw documents in heterogeneous formats undergo lightweight cleansing, canonicalization of field names, and storage in a unified schema. The subsequent annotation stage attaches quantitative quality signals, including structural and linguistic metrics, and applies masking to all personally identifiable information (PII). The filtering stage then materializes stage-specific corpora by applying threshold rules to the annotated data and serializes the result into shard files optimized for streaming.

\shortsection{Data Filtering}  
Korean-specific data filtering schemes have been largely underexplored from the literature. To obtain a corpus that is simultaneously broad and reliably high-quality, we devise a two-tier filtering framework tailored to the linguistic and typographic characteristics of Korean. The first tier extends the rule sets of \citet{weber2024redpajama} and \citet{lozhkov2024fineweb-edu} by redesigning every heuristic for Korean morphology. Among various quantitative signals, five representative examples—symbol-to-word ratio, mean word length, sentence count, masked-PII ratio, and the proportion of normalized to raw length—are computed for each document. Target ranges for these signals are established through manual inspection with an internal reviewer, and thresholds are further adapted to each source domain (e.g., blogs, wikis) to suppress noise while preserving recall. 

The second tier employs model-based scoring. FastText \citep{bag-of-tricks, fasttext} and transformer encoders are trained under two supervision regimes. In the binary regime, wiki-like passages constitute positive examples whereas noisy web pages form the negative class; the posterior probability furnishes a continuous quality score. In the ordinal regime, a language model assigns $0$–$5$ ratings for educational utility, informativeness, and narrative coherence, producing “wiki-like”, “educational”, and “explanatory” quality predictors analogous to GPT-$3$, FineWeb-edu, and DCLM filters \citep{gpt3, fineweb-edu, dclm}. A document is retained only if it satisfies a stage-specific conjunction of heuristic thresholds and model scores. Near-duplicates are removed with a MinHash index that is rebuilt at every refresh. 

Table \ref{tab:stage_data_ratios} summarizes the document‐level yield rates of sub-sampled data achieved by the two‐tier pipeline. Even within this modest slice, the first stage discards roughly $90$ \% of raw pages overall, while the more selective second stage retains just $1$ – $20$ \%. These figures reveal aggressive corpus compression, with the pipeline condensing the raw crawl by roughly one to two orders of magnitude even on the sub-sampled slice.

\shortsection{Synthetic Data Generation} \label{subsec:synthetic_data}
In contrast to the extensive curated resources available for major languages (e.g., English and Chinese), high-quality Korean corpus remains markedly under-represented. To redress this asymmetry, we initiate a systematic program of high-fidelity synthetic data generation, focusing on domains—such as education, law, historical facts, and cultural sentiment—where native Korean content is especially sparse \citep{RFT, corgi}. Leveraging our in-house model family, the pipeline follows two complementary tracks, rewriting existing documents and generating new text from curated seed prompts, while placing filtering and verification at the core of the process to ensure that only high-fidelity Korean data is retained.

\begin{table}[th]
    \centering
    \begin{tabular}{lcc}
    \toprule
    Data & Stage $1$ Yield (Filtered / Raw) & Stage $2$ Yield (Filtered / Raw) \\
    \toprule
    Total & 9.59\% & 1.36\% \\
    \midrule
    Blog & 57.74\% & 19.84\% \\
    Cafe & 31.53\% & 2.35\% \\
    Web & 4.49\% & 0.27\% \\
    \bottomrule
    \end{tabular}
    \caption{Stage-wise document yield rates after two-tier filtering.}
    \label{tab:stage_data_ratios}
\end{table}

The synthetic-data workflow comprises four coupled phases \citep{PIT, textbooks2, benallal2024cosmopedia, nemotron-cc}. (1) Data-design phase: We draft a specification that fixes the target domain, desired volume, file format, and downstream use case. This document governs every subsequent decision in the pipeline. (2) Seed-acquisition and generation phase: License-compliant seed material is collected from open-source and internal repositories. These seeds are either paraphrased to remove copyright artifacts or expanded into new passages through prompt-based generation with our in-house language-model family. (3) Filtering and refinement phase: The resulting text is processed by the same two-tier filtering stack used for web data, augmented by routines that detect repetitive templates, logical inconsistencies, and machine-like phrasing. (4) Integration phase: Only data that satisfy all quality checks are versioned and merged into the pre-training corpus, ensuring that synthetic examples extend coverage without degrading overall corpus fidelity. We provide illustrative synthetic data examples in Appendix \ref{appendix:synthetic_example}. These synthetic corpora are injected into both Stage $1$ and Stage $2$ of the pre-training curriculum.

\subsection{Model Architecture} \label{pre-training:architecture} 
On the architectural front, our design integrates three key components—(i) a compute–memory-balanced Transformer layout \citep{chichilla, gemma2}, (ii) Peri-Layer Normalization (Peri-LN, \citet{peri-ln}) for training stability and performance, and (iii) Maximal Update Parametrization ($\mu$P, \citet{tensorprogram4, tensorprogram6}) for scale-robust hyper-parameter transfer—together enabling stable scaling and cost-efficient training. 

\shortsection{Compute–Memory Balanced Architecture}
To minimize compute-bound training cost and memory-bound inference latency under a fixed parameter budget, we employ a \emph{shallower-but-wider} Transformer configuration \cite{chichilla, gemma2}. The model reduces the number of blocks and reallocates the freed parameters to larger hidden and feed-forward dimensions. Because each self-attention layer incurs $O(L^{2})$ FLOPs and $O(L)$ activation memory with respect to sequence length $L$, lowering depth proportionally decreases attention overhead, while widening the FFN, whose cost grows linearly in $L$, maintains representational capacity. 

To empirically substantiate this design, we start from a $3$B-parameter baseline comprising $26$ Transformer blocks with an FFN hidden size of $7,168$ and generate a shallow-but-wide variant by reducing the depth to $18$ layers ($30$ \% shallower) while proportionally increasing the FFN hidden dimension to $11,264$ ($57$ \% wider), thereby conserving the total parameter budget. Owing to the quadratic attention cost, this reallocation lowers the theoretical compute for an $8$K-token sequence by $13.7$ \% TFLOPs. Consistently with this analysis, the modified model ingested $15$ \% more training tokens within an identical wall-clock budget and matched the validation perplexity of the deeper control, confirming that width-centric capacity reallocation preserves modeling quality while conferring tangible hardware efficiency.

\shortsection{Stability-Oriented Transformer}
We stabilize scale-up by coupling Maximal Update Parametrization ($\mu$P) with a Peri-Layer-Normalized Transformer. Following the $\mu$Transfer procedure, we sweep learning-rate and regularization only on small proxy models, then zero-shot port the optimal settings to each production scale. Because $\mu$P preserves update magnitudes across configurations, the large models inherit well-conditioned gradient norms without further tuning, greatly reducing exploration cost while keeping feature learning intact \citep{tensorprogram4, tensorprogram6}. 

Peri-Layer Normalization (Peri-LN) normalizes both the input and output of every Transformer sub-layer, bounding hidden-state variance to grow at most linearly with depth and that layer-wise gradient norms remain stable throughout training.By tightly bounding hidden state statistics, Peri-LN suppresses the massive activations typically observed in Pre-LN models \citep{massive-activation}. Peri-LN also removes the need for FLOP-intensive ablation studies to stabilize architectural or training hyper-parameters. Empirically, Peri-LN yields lower pre-training loss and smaller run-to-run variance \citep{peri-ln}. Maximal Update Parametrization ($\mu$P) complements Peri-LN by preserving optimization statistics across width and depth, so hyper-parameters tuned on sub-billion-parameter proxies transfer reliably to multi-billion-parameter instances. Together, Peri-LN and $\mu$P provide a principled, cost-effective pathway to stable scaling. 

\begin{figure}[t]
  \centering
  \begin{subfigure}[b]{0.5\textwidth}
    \centering
    \includegraphics[width=\textwidth]{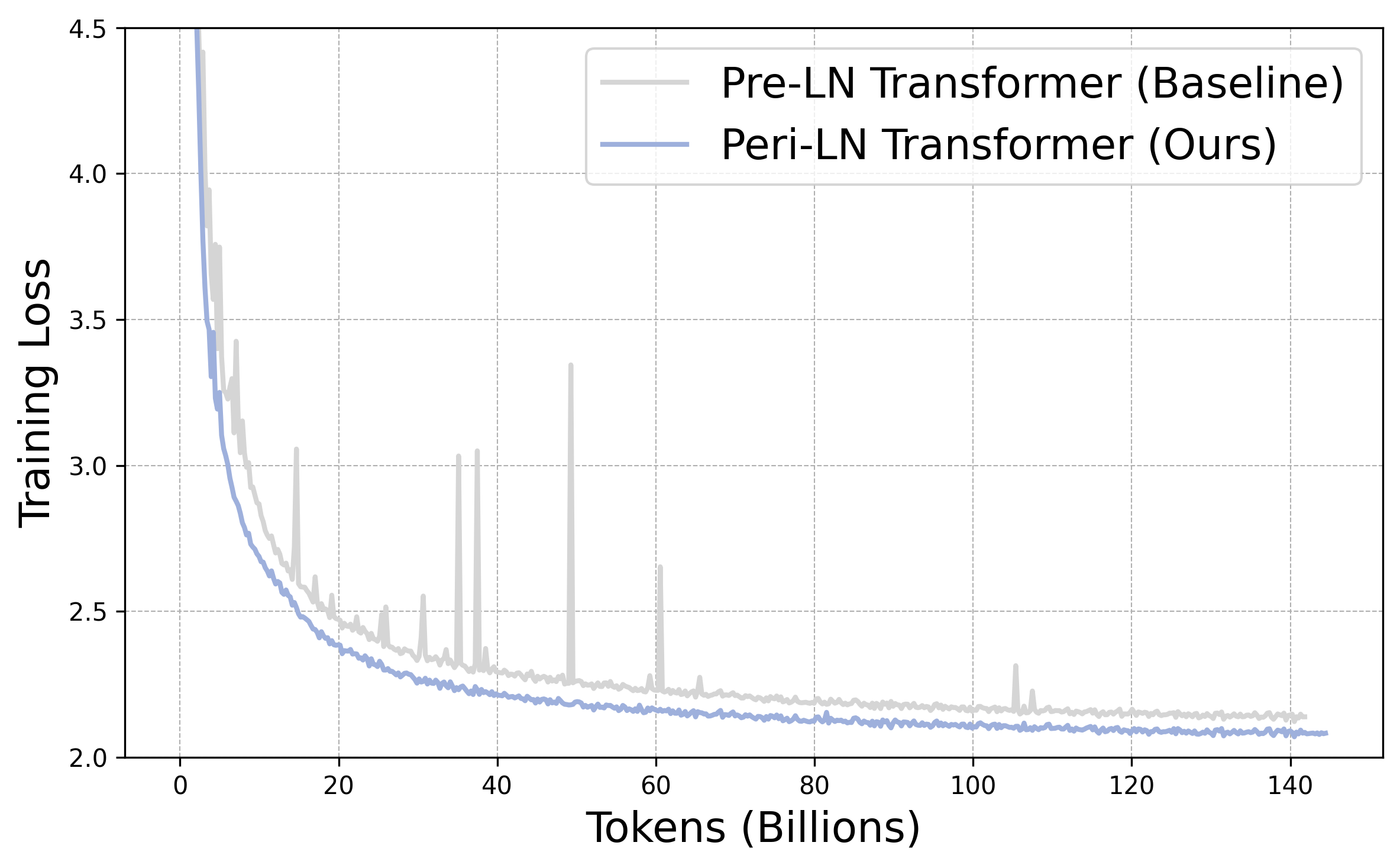}
    \caption{Training loss}
    \label{fig:peri-a-loss}
  \end{subfigure}%
  \hfill
  \begin{subfigure}[b]{0.5\textwidth}
    \centering
    \includegraphics[width=\textwidth]{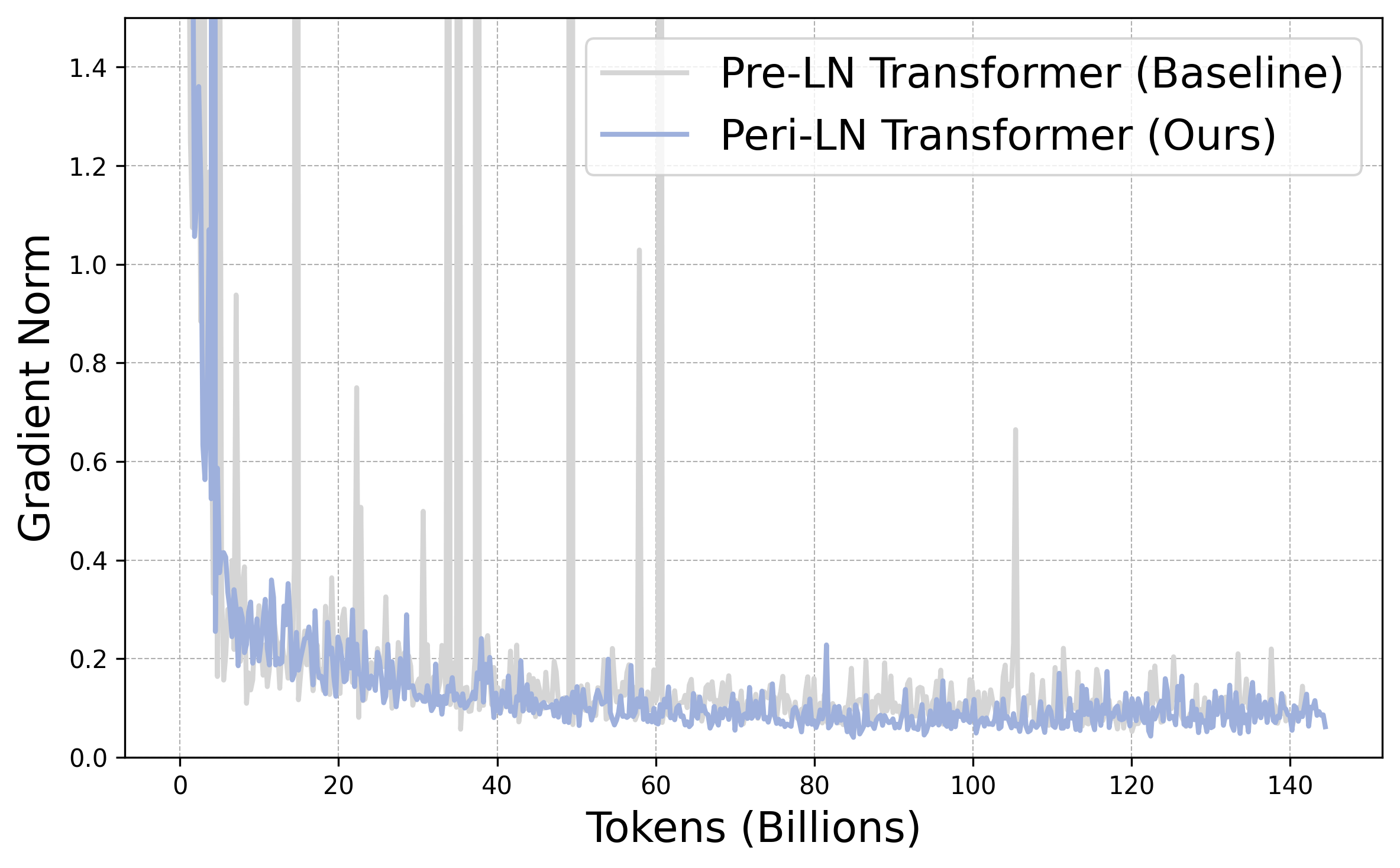}
    \caption{Gradient-norm}
    \label{fig:peri-b-grad}
  \end{subfigure}%
  \caption{Performance comparison between $8$ B-parameter Pre-LN and Peri-LN Transformers during pre-training. Each model size excludes the embedding parameters.}
  \label{fig:peri}
\end{figure}

To evaluate normalization choices at production scale, we trained two Llama-style models \citep{llama3} with $8$ B parameters on the same open-corpus dataset \citep{nemotron-cc}, along with our in-house version of the TikToken tokenizer\footnote{\url{https://github.com/openai/tiktoken}}: a standard Pre-LN \citep{onlayer} baseline and an otherwise identical Peri-LN variant. As illustrated in Figure \ref{fig:peri}, the Peri-LN model exhibits fewer gradient and loss spikes than its Pre-LN counterpart, reproducing the large-scale stability benefits reported by \citet{peri-ln}. Furthermore, the Peri-LN configuration attains, on average, a $15$ \% lower training loss within the same wall-clock budget. These findings confirm that Peri-LN delivers superior stability and performance without incurring additional computational cost, and thus we adopt it as the default normalization scheme in the \hcxt architecture.

\subsection{Pre-Training Curriculum}   \label{pre-training:stage} 
We adopt a three-staged pre-training curriculum, with each phase focused on a distinct capability target \citep{olmo2, minicpm}. Stage $1$ establishes a general-purpose foundational knowledge base. Stage $2$ refines domain-specialized competence by continuing training on high-quality corpora. Stage $3$ extends the context window to $128$K tokens and internalizes long chain-of-thought reasoning by fine-tuning on rejection-sampled traces generated from an in-house model family. The staged curriculum strategically allocates computational FLOPs to phases with the highest marginal utility, optimizing cost-efficiency while maximizing incremental performance gains.

\shortsection{Stage 1: Foundational Knowledge Construction}
The first training stage establishes a broad knowledge base spanning multiple domains. We curate a multilingual corpus, principally Korean and English. Training proceeds on sequences up to $8$K tokens, consuming $6$ trillion tokens in total. The learning rate is linearly increased during the initial $5,000$ steps to a peak of $1.59e$-$3$ determined by $\mu$P scaling, after which it is annealed according to a cosine schedule to $1.59e$-$4$ ($10$ \% of the maximum), thereby promoting stable convergence.

\shortsection{Stage 2: Domain-Specialized Capability Boosting}
The mid-training stage introduces an additional 1 trillion tokens to sharpen the model's domain expertise and reasoning ability while maintaining the $8$K-token context length established in Stage $1$. We gradually down-weight generic web text and increase high-quality, domain-focused corpora including the synthetic datasets constructed in Section~\ref{subsec:synthetic_data}. A brief $2,000$-step warm-up ensures a smooth transition to these revised distribution. 

Guided by \citet{deepseek-llm}, for learning rate schedule, we adopt a two-step decay profile: the rate is held at $1.59e$-$4$ for $80$ \% of training, reduced to $31.6$ \% of this peak ($\approx 4.76e$-$5$) for the next $10$ \%, and finally to $10$ \% ($\approx 1.59e$-$5$) for the last $10$ \%. For the data mix, following \citet{does-your-data-spark-joy}, we rebalancing the dataset during the final $10$ \% of training steps. Sampling of lower-quality general text is gradually reduced. Conversely, the sampling weight of under-represented domains, crucial for sovereign-AI applications, is increased, with emphasis on Korean medical literature, national economic reports, and culturally contextualized historical archives.

\shortsection{Stage 3: Extended Context Alignment}
Standard corpora are biased toward short documents; naively over-sampling longer texts therefore disrupts training stability \citep{scaling-long-context}. We mitigate this issue with \emph{length-based, proportion-preserving resampling}, which increases the number of long documents while maintaining each length bucket’s share of total tokens. After pre-training with an $8$ K context window and a rotary-position-embedding base $\theta$ of $500$ K, we expand the window in three successive stages—$32$ K, $64$ K, and $128$ K. At each expansion, $\theta$ is raised from $500$ K to $5$ M, then to $20$ M, and finally to $100$ M. A brief warm-up followed by cosine decay restores perplexity before the next enlargement \citep{roformer, rope}. To supply explicit supervision for extended reasoning, we additionally train on a long chain-of-thought corpus generated in-house and filtered via rejection sampling \citep{RFT, corgi} (see \S\ref{subsec:synthetic_data}). This synthetic dataset spans up to $128$ K tokens, enabling the model to master long-context conditioning without degrading the general or domain-specific competencies obtained in Stages $1$ and $2$.

\section{Post-Training}
\begin{figure}[tbp]
    \centering
    \includegraphics[width=\textwidth]{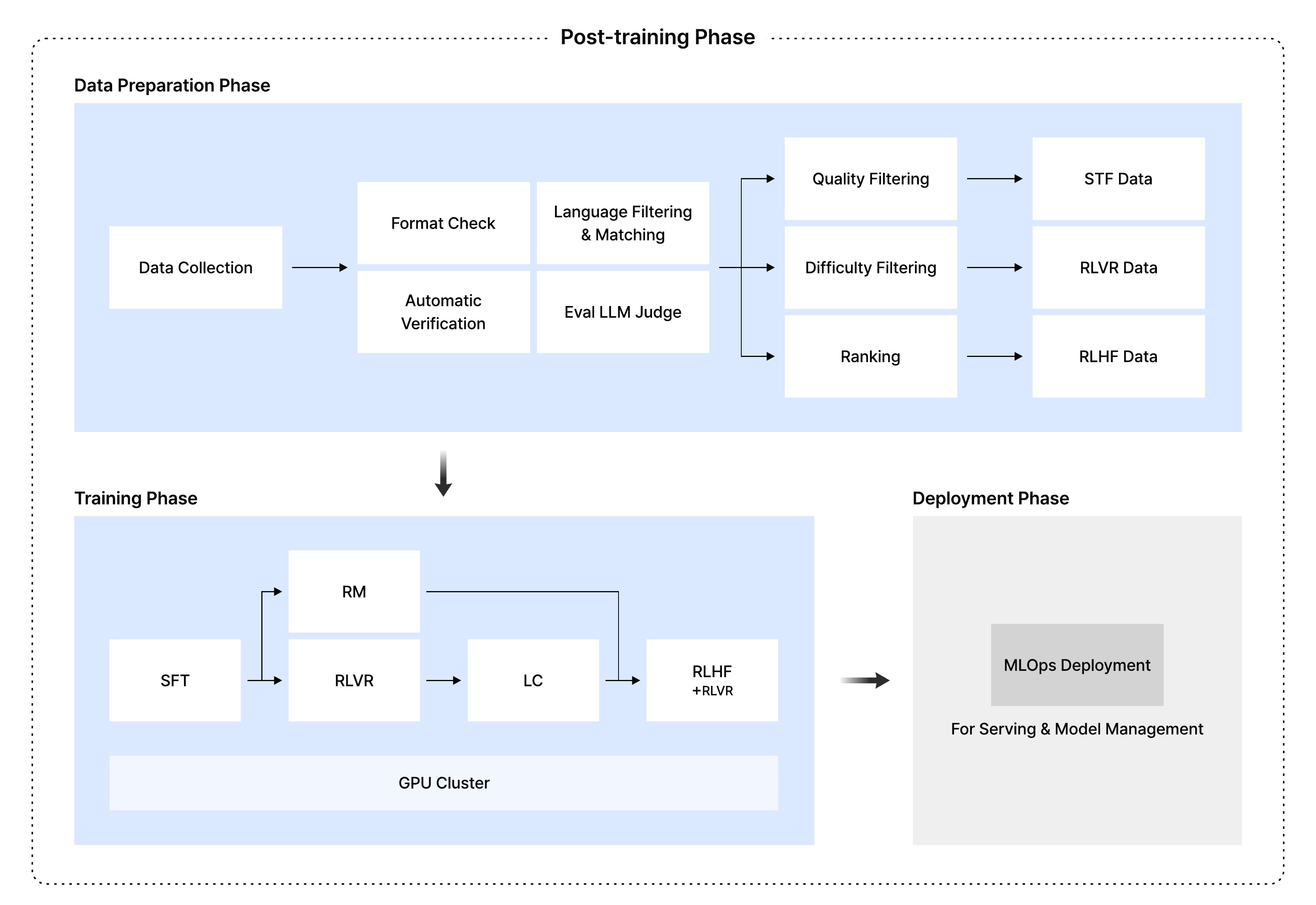}
\caption{Post-training pipeline of \hcxtfull. (1) Data-Preparation Phase: Data is collected and then rigorously processed through steps such as format validation, automatic verification, language-based filtering, and evaluation by LLM-based judges. The data is refined through quality filtering, difficulty filtering, and ranking to prepare data suitable for subsequent training stages with different objectives. (2) Training Phase: A sequence of fine-tuning procedures—including Supervised Fine-Tuning (SFT), Reward Modeling (RM), Reinforcement Learning with Verifiable Rewards (RLVR), training for reasoning Length Controllability (LC), and Reinforcement Learning from Human Feedback (RLHF) —is executed across a large-scale GPU cluster.}
    \label{fig:overview_post_training}
\end{figure}

This section outlines the post-training methodology of THINK: a supervised fine-tuning (SFT) phase that injects core reasoning patterns and task-specific capabilities (Section \ref{text-sft}); and a multi-stage reinforcement learning pipeline that incorporates verifiable rewards, length-controllability, and human feedback to achieve aligned, efficient, and scalable reasoning (Sections \ref{text-rlvr}–\ref{text-rlhf}). See Figure \ref{fig:overview_post_training} for an overview of the training process.

\label{reasoning-mode}

\begin{table}[ht!]
\centering
\vspace{0.5em}
\small
{
\renewcommand{\arraystretch}{1.2} 
\begin{tabular}{m{0.45\textwidth}|m{0.45\textwidth}}
\hline
\textbf{\rule{0pt}{1.2em}Reasoning Mode} & \textbf{\rule{0pt}{1.2em}Non-Reasoning Mode} \\
\hline
\begin{minipage}[t]{\linewidth}\ttfamily
\vspace{0.5em}
\texttt{<|im\_start|>user}\\
\textcolor{blue}{\texttt{\{query\}}}\texttt{<|im\_end|>}\\
\texttt{<|im\_start|>assistant}\textcolor{red}{\texttt{/think}}\\
\textcolor{blue}{\texttt{\{reasoning\}}}\texttt{<|im\_end|>}\\
\texttt{<|im\_start|>assistant}\\
\textcolor{blue}{\texttt{\{response\}}}\texttt{<|im\_end|><|endofturn|>}
\vspace{0.5em}
\end{minipage}
&
\begin{minipage}[t]{\linewidth}\ttfamily
\vspace{0.5em}
\texttt{<|im\_start|>user}\\
\textcolor{blue}{\texttt{\{query\}}}\texttt{<|im\_end|>}\\
\texttt{<|im\_start|>assistant}\\
\textcolor{blue}{\texttt{\{response\}}}\texttt{<|im\_end|><|endofturn|>}
\vspace{0.5em}
\end{minipage}
\\
\hline
\end{tabular}
}
\caption{Unified chat template used for training models to support both reasoning and non-reasoning interaction modes.}
\label{tbl:chat-template}
\end{table}

\hcxt is trained to operate in an integrated manner, allowing for dynamic switching between a detailed `reasoning mode' for complex, multi-step reasoning and a more direct `non-reasoning mode' for rapid, context-driven responses. This unified framework eliminates the need for users to switch between separate models (e.g., a dedicated reasoning model and a chatbot), as illustrated in Table~\ref{tbl:chat-template}.

\subsection{Supervised Fine-Tuning (SFT)} \label{text-sft}

Supervised Fine-Tuning (SFT) serves as a foundational step in our post-training pipeline, aiming to inject desired behaviors and reasoning patterns into the model. This stage establishes a strong base for subsequent reinforcement learning phases.

\begin{figure}[tbp]
    \centering
    \includegraphics[width=\textwidth]{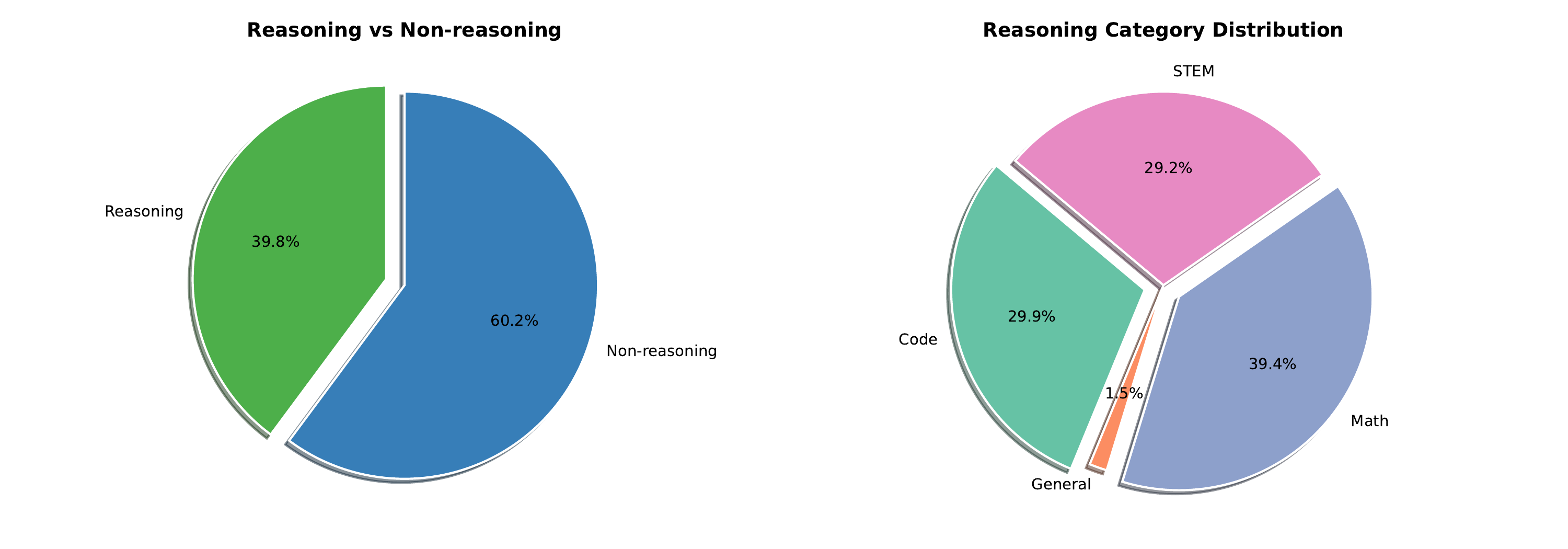}
    \caption{Data distribution utilized for Supervised Fine-Tuning (SFT), reflecting a balanced composition tailored to support effective downstream reinforcement learning and reasoning capabilities.}
    \label{fig:distribution}
\end{figure}

The dataset used for SFT is constructed by aggregating various sources across mathematics, coding, STEM, and general abilities. We carefully curate data from a series of ablation studies and utilize high-quality open-source and in-house data. For reasoning data, each sample contains prompt, assistant think, and assistant response. The assistant think contains a rather free-form chain-of-reasoning, while the assistant response is a concise, finalized output that directly answers the user's query based on that reasoning. The general statistics for the SFT dataset is illustrated in Figure \ref{fig:distribution}.

To ensure data quality and consistency, we apply a multi-stage filtering pipeline across all datasets. Each item in data goes through a basic format check to ensure that the output contains proper format (e.g., boxed answers for math problems and compilability for code problems). Language filtering is applied to select only samples written in the target language, and language matching further ensures that input and output languages are the same for each sample. For reasoning data specifically, we also check whether the final answer is automatically verifiable. For non-reasoning data, we employ a LLM-as-a-Judge method to score each example by their helpfulness and safety and filter out those with low scores.

Training is performed with dynamic batching to fill each batch dynamically to its maximum capability, in order to optimize GPU utilization and memory usage. The model is trained over 4 epochs with early stopping based on validation accuracy. Similarly to other reports \citep{yang2025qwen3technicalreport}, we observe that selecting a checkpoint from later epochs results in reduced exploration of the model during the subsequent phase. More comprehensive details on the training setup of SFT are provided in our previous technical report \citep{yoo2024hyperclovaxtechnicalreport}. 

\subsection{Reinforcement Learning with Verifiable Rewards (RLVR)} \label{text-rlvr}

The Reinforcement Learning with Verifiable Rewards (RLVR) is crucial for improving reasoning capabilities through verifiable feedback mechanisms. The main objective is to optimize model performance by accurately guiding behavior through precise rewards and penalties. 

\shortsection{Reinforcement Learning Algorithm} \label{rlvr_algorithm}
In our implementation of RLVR, we adopt Group Relative Policy Optimization (GRPO) \citep{shao2024deepseekmath}. 

Unlike more traditional RL algorithms, 
it calculates a baseline advantage based on multiple generations per prompt, optimizing computational efficiency and maintaining training effectiveness.
To further enhance the robustness and accuracy of our RLVR framework, we introduce several targeted modifications:

\begin{itemize}
    \item \textbf{KL Divergence Penalty Removal:} Our initial experiments indicated that this penalty restricts models from exploring diverse behaviors and incurs significant computational overhead due to the necessity of inference from a reference model. Removing the penalty improved computational efficiency and model flexibility.
    \item \textbf{Constant Normalization:} We observed that prompt difficulty often correlates with response length–more difficult prompts tend to produce longer responses–thereby introducing biases related to response length. To mitigate these biases from varying response lengths and prompt difficulties, we adopt constant normalization strategy from \citet{liu2025understanding}.
    \item \textbf{Relaxed Upper Bound for Exploration:} To encourage exploration and prevent deterministic policy collapse, we adopt the clip-higher approach \citep{yu2025dapo}, which raises the upper threshold of the importance sampling ratio in GRPO. By including low-probability tokens into policy updates, this approach increases policy entropy and fosters diverse reasoning paths.
\end{itemize}

Collectively, these methodological enhancements enable our RL training to achieve an optimal balance of exploration, computational efficiency, and stable training performance.

\shortsection{Data Efficiency}
To optimize training efficiency, enhance model performance, and effectively utilize computational resources, we employ targeted difficulty filtering techniques, including both offline and online methods.

We implement offline difficulty filtering to our dataset by excluding prompts that are either too easy or too challenging. Specifically, we leverage predictions generated by the SFT checkpoint---our initial model for RLVR---to evaluate the difficulty of each prompt. By sampling multiple responses from this checkpoint, we calculate the average accuracy of predictions and remove prompts with accuracy of exactly 0.0 or 1.0. This strategy ensures the inclusion of prompts only with appropriate difficulty levels at the outset of training.

However, offline difficulty filtering has limitations. Because this filtering method occurs only once before the training begins, it is inherently static. As the model's performance improves as the training progresses, the dataset's difficulty level cannot be adjusted accordingly–a problem that was once challenging can become solvable. Consequently, this static nature can lead to discrepancies between evolving model capabilities and fixed difficulty of prompts.

To address the shortcomings of offline filtering, we additionally incorporate an online difficulty filtering strategy. Utilizing GRPO allows us to generate multiple responses per prompt within each batch. For each group, we calculate accuracy and remove prompts where all generated responses are either entirely correct or entirely incorrect from the batch. This dynamic filtering approach continuously adapts the training set’s difficulty to the model's evolving capabilities, ensuring that learning remains focused on informative examples and thereby maintaining optimal training efficiency.

Our analysis aligns with recent findings suggesting that online difficulty filtering effectively optimizes the lower bound learnability of reinforcement learning algorithms by dynamically balancing prompt difficulty \citep{bae2025online}. Importantly, we observe that even with initial offline filtering, online filtering still provides substantial additional benefits. Thus, combining both offline and online difficulty filtering significantly enhances our training efficiency and model performance.

\shortsection{Reward Shaping}
To effectively guide model training and enhance its performance in our RLVR framework, we carefully design a reward shaping strategy consisting of several distinct components:

\begin{itemize}
    \item \textbf{Format Reward:} We establish a set of format rules that responses must follow. To calculate this reward, we count the number of rules adhered to by the model's response and divide it by the total number of format rules.
    We adopt this soft penalty approach as it demonstrates minimal negative impact on reasoning performance, allowing models to progressively align with the desired response structure without detrimental effects.
    
    \item \textbf{Language Reward:} This reward is computed based on the ratio of characters generated in the same language as the prompt. By directly correlating the language of responses with the language of prompts, this reward encourages the model to reason in the intended language, significantly enhancing multilingual reasoning capabilities.

    \item \textbf{Verifiable Reward:} We incorporate verifiable rewards across multiple problem categories, including mathematics, code generation, code input-output (Code IO), and multiple-choice questions. The verification outcomes directly determine reward allocation, with a binary value: a fully correct response receives a reward of 1.0, while any incorrect response results in a reward of 0.0.

    \item \textbf{Overlong Reward:} We adopt both Soft Overlong Penalty and Overlong Loss Masking \citep{yu2025dapo}, because penalizing truncated samples harshly can introduce undesirable reward noise, potentially destabilizing training by penalizing valid reasoning solely due to length. 
    The former gradually increases as the response length exceeds the predefined maximum value, and the latter masks the loss of truncated samples, effectively stabilizing the training process. 
\end{itemize}

\shortsection{Optimized Rollout Sampling Process}
Efficiency in the rollout sampling process is crucial for optimizing the RLVR training pipeline, as this stage typically dominates the overall training duration. To address this, we implement a highly efficient asynchronous sampling procedure. In this setup, inference nodes are utilized continuously and concurrently until the number of completed rollout samples meets or exceeds the training batch size. Samples generated from these inference nodes are collected and stacked asynchronously, significantly reducing idle times and improving resource utilization.

Moreover, due to our implementation of online difficulty filtering, certain samples may be dynamically filtered out during the rollout process, potentially causing delays or inefficiencies. To counteract this, we maintain a buffered approach to concurrent sampling, ensuring multiple samples are processed simultaneously. This strategy effectively compensates for any filtered-out examples by ensuring continuous generation of alternative samples, thereby minimizing or entirely masking the time loss associated with discarded examples. This optimized asynchronous sampling approach greatly enhances the efficiency and stability of the RLVR training process \citep{bae2025online}.

\subsection{Reasoning Length Controllability (LC)}

Reinforcement learning with Large Reasoning Models (LRMs) enables drastic improvements in complex reasoning capabilities, but often accompanies undesired tendencies to overthink \citep{chen2024not, sui2025stop} or even underthink \citep{wang2025thoughts} compared to the optimal reasoning length. For practical and flexible deployment of computationally expensive LRMs, we identify \emph{length controllability} (LC) as a key desideratum. To induce LC in HyperCLOVA X THINK, we additionally incorporate the length-penalized reward functions introduced by \citealp{aggarwal2025l1}.

On top of the training configurations from the previous RLVR stage, we train our model on the length-penalized reward functions (L1-Exact and L1-Max) from \citealp{aggarwal2025l1}. We append `\texttt{Think for maximum N tokens}' on the input instructions, where we sample N from a discrete token budget set of $\mathcal{B} = \{1024, 2048, 4096, 8192, 16384\}$ to accelerate LC capability\footnote{The original L1 paper randomly samples N from $\mathcal {U}_{[100,4000]}$}. We first train the model on the L1-Exact penalty for about 300 steps to acquire LC and subsequently about 100 steps on the L1-Max penalty to greedily reduce the reasoning length when possible.

\subsection{Reinforcement Learning from Human Feedback (RLHF)} \label{text-rlhf}
Reinforcement Learning from Human Feedback (RLHF) aligns model outputs with human preferences and practical usability. By combining reasoning/non-reasoning RLHF and RLVR, we concurrently refine model behavior to improve alignment with human preferences while preserving and enhancing reasoning abilities.

To better align the model’s outputs with human preferences, we first train a reward model using a combined set of human preference data, as detailed in our previous technical report \citep{yoo2024hyperclovaxtechnicalreport}. This data consists of pairwise comparisons either annotated by expert raters or inferred via scoring from in-house judge models. The reward model learns to predict scores for each sequence in non-reasoning data. Following this, we use GRPO explained in Section \ref{rlvr_algorithm} as the core RLHF algorithm. The policy is optimized to maximize the expected reward predicted by the reward model. Unlike RLVR, we apply a KL penalty of 0.1 to maintain proximity to the SFT checkpoint. This relatively strong KL penalty prioritizes training stability over exploration in RLHF.

The prompts used during RLHF training consist of a mixture of reasoning and non-reasoning tasks. For non-reasoning, the model is expected to generate assistant response directly, while for reasoning, the model first generates intermediate think step followed by assistant response. The reward model evaluates only the response portion of the output and the think portion is not directly scored, allowing the model to freely develop internal reasoning patterns.

Lastly, when training with RLHF subsequently after RLVR, we observe a slight degradation in the model’s reasoning ability that was optimized during the RLVR phase. A similar pattern was also observed in other reasoning models \citep{yang2025qwen3technicalreport}. To address this issue, we adopt a joint training strategy where RLVR and RLHF are trained concurrently. Specifically, we interleave the training batches such that each batch contains a mixture of samples from RLVR and RLHF datasets. This approach preserves the performance gains of both RLHF and RLVR while unifying the training phases, resulting in a simpler and more effective training pipeline.

\section{Evaluation}
\subsection{Baselines}~\label{sec:baselines}
We compare our model against publicly available models of comparable size that are recognized for their reasoning capabilities, including Qwen3-14B, Qwen3-32B~\citep{yang2025qwen3technicalreport}, QwQ-32B~\citep{qwq32b}, and EXAONE-Deep-32B~\citep{exaone-deep}. 
We utilize evaluation scores directly from each model's original paper when available. Otherwise, we conduct our own evaluations and report the corresponding results.

\subsection{Evaluation Protocol}\label{subsec:caliper}
When published metrics are unavailable, we perform in-house evaluations using primarily public benchmark sets, with the exception of KoBigBench \citep{yoo2024hyperclovaxtechnicalreport}. The primary goal of our evaluation strategy is to interpret and extract the predicted answers from language models for both open-ended and multiple-choice benchmarks as accurately as possible. Models often fail to produce a final answer when asked to generate the reasoning chain and the answer consecutively in a single pass. To address this, we adopt a two-pass generation scheme: the model first produces the reasoning chain with our chat template (\verb=<|im_start|>assistant/think\n...<|im_end|>=), and we then generate the answer by appending an answer prefix. Our evaluation framework combines LM Eval Harness \cite{eval-harness} with an in-house toolkit that we plan to release soon for public reference. 

For our model, we configure the generation temperature at 0.5 and top-p at 0.95. In the case of other models, their authors' recommended optimal hyperparameters are utilized. All evaluations are performed using zero-shot Chain-of-Thought (CoT) reasoning, and the maximum CoT generation length is uniformly set to 4096.
    
\subsection{Korea-Centric Benchmarks}~\label{sec:core_bench}
\hspace{-.3em}\shortsection{Setup}
    As introduced in Section 1, our model's general performance is evaluated against various baselines using a set of Korea-centric benchmarks. These evaluations are designed to assess the model's understanding of Korean culture and knowledge. To achieve this, we curated datasets specifically pertaining to Korea:

    \begin{itemize}
        \item \textbf{General Aptitude:} 
        KMMLU~\citep{son-etal-2025-kmmlu} and CSAT gauge general Korean knowledge. KorMedMCQA~\citep{kweon2024kormedmcqamultichoicequestionanswering} focuses on medical problem-solving and KoBALT-700~\citep{shin2025kobaltkoreanbenchmarkadvanced} assesses linguistic depth and typological grounding in Korean.
        \item \textbf{Culture and Language:} 
        HAERAE-1.0~\citep{son-etal-2024-hae},  CLIcK~\citep{kim-etal-2024-click}, and KoBigBench\footnote{The dataset will be publicly released.} evaluate Korean-specific cultural, geographical, historical knowledge, etc.
        \item \textbf{Instruction-Following:} LogicKor~\citep{logickor} and KoMTBench~\citep{KoMT-Bench} measure the model's ability to follow Korean instructions.
    \end{itemize}

\begin{figure}[tbp]
    \centering
    \includegraphics[width=1.0\textwidth]{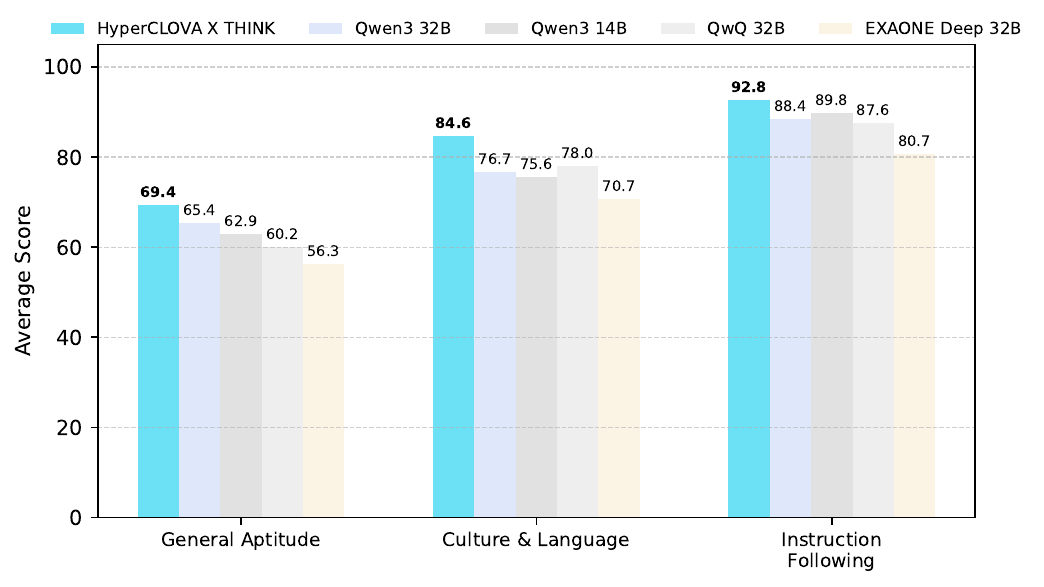}
    \caption{Summary of model performance on (1) General Aptitude, (2) Culture and Language, and (3) Instruction-following benchmarks specifically focused on Korea. The instruction-following benchmark scores are normalized by multiplying their original values by 10.}
    \label{fig:overall_score}
\end{figure}

\shortsection{Result} 
    Our model's strong performance on the comprehensive aptittude tests, Korean-specific culture and linguistic benchmarks, and a suite of benchmarks for probing instruction-following capabilities is summarized in Figure~\ref{fig:overall_score} and detailed in Table~\ref{tab:comprehensive_korea_benchmark}. By employing zero-shot CoT prompting to elicit robust reasoning and evaluating answers based on accuracy, we demonstrate that THINK surpasses other baselines. Additional evaluation results can be found in the Appendix~\ref{appendix:math_coding_benchmark} and Appendix~\ref{appendix:text-only-analysis}. Furthermore, this superior performance is achieved with a relatively small computational cost, which will be discussed further in the subsequent section.

\begin{table*}[t]\centering\small
\renewcommand{\arraystretch}{1.6}%
\resizebox{\textwidth}{!}{%
\begin{tabular}{p{1.4cm}p{2cm}>{\columncolor{customblue}\centering\arraybackslash}p{1.5cm}
>{\centering\arraybackslash}p{1.5cm}
>{\centering\arraybackslash}p{1.5cm}
>{\centering\arraybackslash}p{1.5cm}
>{\centering\arraybackslash}p{1.5cm}}
\toprule
\textbf{Category} & \textbf{Benchmarks} & \textbf{HCX THINK} & \textbf{Qwen3} & \textbf{Qwen3} & \textbf{QwQ} & \textbf{EXAONE Deep} \\
 & & ( - ) & (32B) & (14B) & (32B) & (32B) \\
\midrule
\multirow{4}{*}{\makecell{General \\ Aptitude}} & \textbf{KMMLU} & \textbf{69.7} & 63.5 & 49.3 & 54.1 & 53.6 \\
 & \textbf{CSAT} & 83.2 & 81.9 & 77.1 & \textbf{84.7} & 69.7 \\ 
 & \textbf{KorMedMCQA} & \textbf{76.0} & 74.7 & 68.5 & 69.4 & 68.8 \\ 
 & \textbf{KoBALT} & \textbf{48.9} & 41.4 & 38.4 & 32.4 & 33.0 \\
\midrule
\multirow{3}{*}{\makecell{Culture \& \\ Language}} & \textbf{HAERAE} & \textbf{87.8} & 75.1 & 74.1 & 76.2 & 74.7 \\
 & \textbf{CLIcK} & \textbf{80.1} & 71.1 & 68.8 & 73.6 & 62.2 \\
 & \textbf{KoBigBench} & \textbf{85.9} & 83.9 & 83.8 & 84.1 & 75.3 \\
\midrule
\multirow{2}{*}{\makecell{Instruction-\\Following}} & \textbf{LogicKor} & \textbf{9.65} & 8.93 & 9.15 & 9.02 & 8.54 \\
 & \textbf{KoMTBench} & \textbf{8.90} & 8.75 & 8.82 & 8.50 & 7.59 \\
\bottomrule
\end{tabular}%
} 
\caption{Performance comparison of language models on Korea-centric benchmarks. Models are evaluated across comprehensive understanding, cultural sovereignty, and chat-based instruction-following tasks, highlighting their capabilities and adaptability within a Korean context.}
\label{tab:comprehensive_korea_benchmark}
\end{table*}

\section{Analysis}

\subsection{Training Efficiency}\label{subsect:training_efficiency}

\begin{figure}[tbp!]
    \centering
    \includegraphics[width=.7\textwidth]{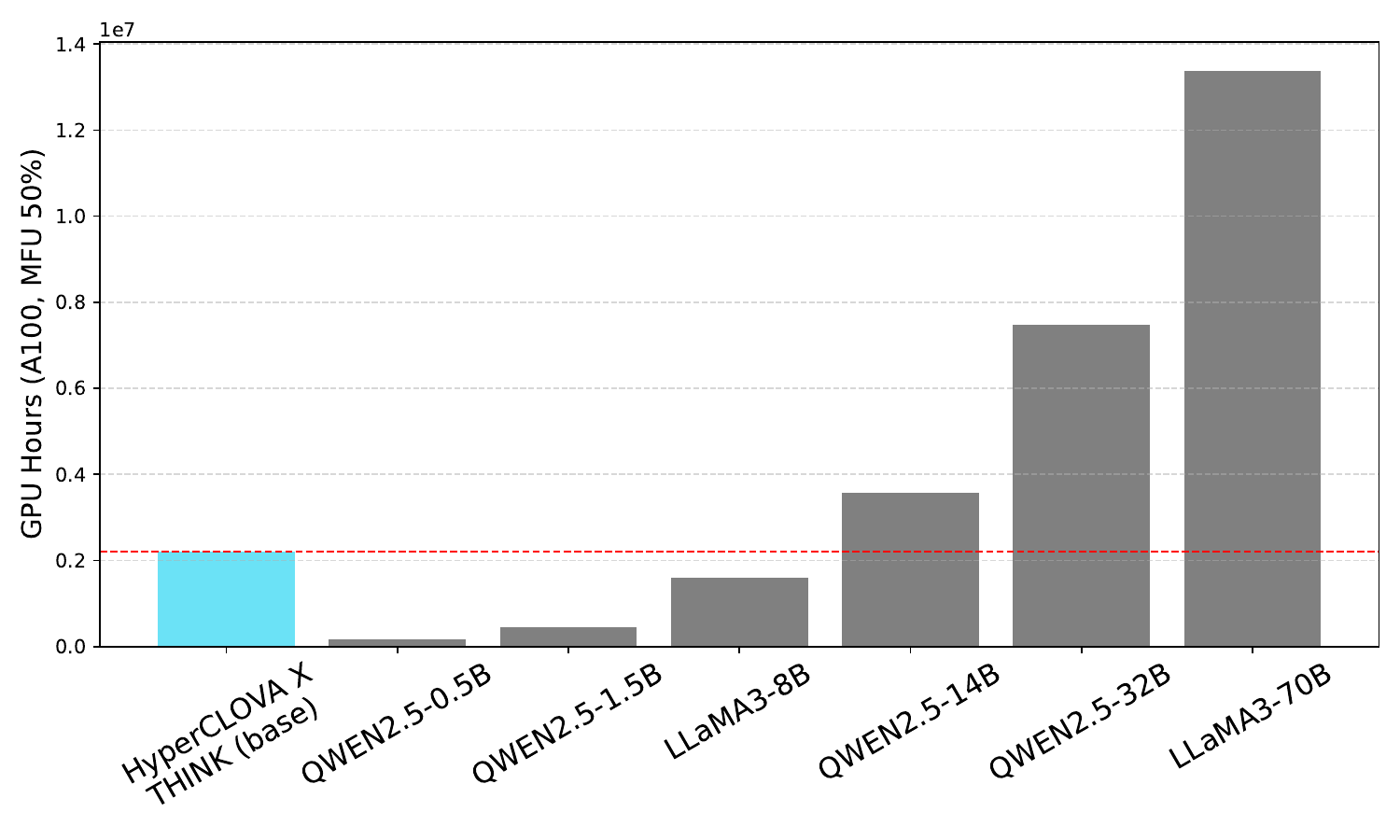}
    \caption{Training Efficiency (GPU Hours / A100 / MFU 50\%)}
    \label{fig:training_efficiency}
\end{figure}

There has been research showing that model performance consistently improves with increases in data volume, parameter count, and computational resources in accordance with Scaling Laws \citep{kaplan2020scalinglawsneurallanguage}. This has also been further supported by more recent work on Expanded Neural Scaling\citep{chang-etal-2024-scaling}. However, recent studies have increasingly emphasized the importance of data quality. For example, \citet{chang-etal-2024-scaling} quantifies data diversity and quality through the concept of effective training tokens and proposes a corresponding scaling law. This study experimentally demonstrates that efficient training and performance improvements are achievable even for smaller models.

It was reported that Qwen2.5 significantly improved its reasoning and long-context generation capabilities using $18$ trillion tokens of high-quality training data and advanced post-training strategies\citep{qwen2025qwen25technicalreport}. Similarly, LLaMA 3, trained on 15 trillion tokens, showed continued performance gains even after surpassing the Chinchilla-optimal range. This suggests that while data scaling remains important, an approach centered on data quality is also necessary. Furthermore, as the volume of natural language data available for collection from the internet approaches its limits, a paradigm shift from data quantity to data quality is accelerating.

\hcxt is developed with a focus on creating a high-efficiency architecture and a training strategy grounded in high-quality data. 
As a result, it required significantly fewer GPU hours than similar sized models to be trained, as shown in Figure~\ref{fig:training_efficiency}.
At the same time, it achieves competitive performance. This demonstrates that strategic data curation and training efficiency are critical factors in developing high-performance LLMs, moving beyond reliance on sheer resource input.

\subsection{Cross-Lingual Transferability}
In this subsection, we investigate cross-lingual consistency and bi-directional translation quality between Korean and English to evaluate whether the model properly transfers the acquired English knowledge into Korean. Our cross-lingual evaluation hypothesis is twofold. First, a model that has efficiently encoded both languages should yield semantically equivalent answers when parallel questions are posed in English and Korean. Second, if the same underlying representations truly capture language-agnostic meaning, the model should also display strong translation ability in both directions. 
    
\shortsection{Cross-lingual Consistency} In order to compute the score of cross-lingual consistency, we adopt the pipeline proposed by~\citet{qi-etal-2023-cross,xing2024evaluating,yoo2024code} and evaluate with Global-MMLU-Lite~\citep{singh2024globalmmluunderstandingaddressing}. We only computed the scores in culturally agnostic samples to exclude examples whose gold answers depend on the source language. Our experiment utilizes Caliper framework, described in Section.~\ref{subsec:caliper}. We categorize the model's predictions on parallel English-Korean MCQA prediction results into four cases: (1) (\ding{51} ,\ding{51}) represents question answered correctly in both languages indicating the desired cross-lingual aligned, (2) (\ding{51}, \ding{55}) is the number of samples that are correct in English but incorrect in Korean, isolating failures of knowledge transfer, (3) (\ding{55}, \ding{51}) represents the opposite scenario, and (4) (\ding{55}, \ding{55}) records items answered incorrectly in both languages, reflecting residual knowledge gaps. This decomposition enables us to attribute improvements in overall accuracy to genuine bilingual robustness.

As shown in Table~\ref{tab:cross_consistency_table}, \hcxt achieves a comparable (\ding{51},\ding{51}) ratio ($74.5\%$) ,only a few points behind the extensively trained Qwen3 32B, while limiting asymmetric errors to 16.5\%. Given that \hcxt was tuned almost exclusively on the Korean–English pair and required a fraction of the compute budget demonstrated in Section ~\ref{subsect:training_efficiency}, this result indicates that carefully targeted bilingual training can offset much of the scale advantage by large multilingual models. Although the remaining asymmetric cases highlight room for improvement, \hcxt already delivers a robust and cost‑efficient bilingual representation. Its overall consistency is also higher than that of EXAONE Deep 32B, suggesting that strategic data curation can sometimes outweigh pure parameter count.

\begin{table}[tb!]
    \centering

    \begin{tabular}{l|rrrr|rr}
        \toprule
        & \multicolumn{4}{c|}{\textbf{Cross-lingual Consistency (En, Ko)}} & \multicolumn{2}{c}{\textbf{MT}} \\ \cmidrule(l){2-7}
        & (\ding{51}, \ding{51}) $\uparrow$ & (\ding{51}, \ding{55}) $\downarrow$ & (\ding{55}, \ding{51}) $\downarrow$ & (\ding{55}, \ding{55}) $\uparrow$& \textbf{Ko$\rightarrow$En} & \textbf{En$\rightarrow$Ko} \\ \midrule
        
        Qwen3 32B & \textbf{81.0} & \textbf{8.0} & \textbf{4.5} & 6.5 & \textbf{92.8}&85.3 \\
        
          EXAONE Deep 32B & 62.5  & 22 &  3.5& \textbf{12.0} & 85.6 & 77.5 \\
          \midrule
        \textbf{HyperCLOVA X THINK} & 74.5 & 12.0 &  \textbf{4.5}&  9.0& 90.3 & \textbf{85.8}\\ \bottomrule
    \end{tabular}
    \caption{Cross-lingual transferability between English and Korean. Each consistency column shows the case of MCQA items for which the model is correct (\ding{51}) or incorrect (\ding{55}) in English (first symbol) and Korean (second symbol). Higher symmetric ((\ding{51}, \ding{51}) and (\ding{55}, \ding{55})) and lower asymmetric ( (\ding{55}, \ding{51}), (\ding{51}, \ding{55})) ratios imply stronger consistency. We also report xCOMET translation quality of Flores on both directions. }  
    \label{tab:cross_consistency_table}
\end{table}

\shortsection{Machine translation} 
To complement the cross-lingual transferability analysis in cross-lingual consistency with MCQA task, we next assess bidirectional machine translation (MT) performance of each model between Korean and English. We adopt the Flores benchmark~\cite{nllb-22} and translate the official sub-samples of test dataset in both directions ( En$\rightarrow$Ko, Ko$\rightarrow$En). A prompt of each model includes the same 1-shot example. We only extract the translation part from the response. Each translation quality is measured with xCOMET-XL~\cite{guerreiro2023xcomet}, a model based metric that has a stronger correlation with professional human judgment compared to BLEU and ChrF. We report xCOMET-XL score for each direction. This performance indicates the model can faithfully re-express the same underlying knowledge as fluent Korean or English. 

MT columns of Table~\ref{tab:cross_consistency_table} provide additional evidence of \hcxt's robust cross-lingual transferability in a generation task. In Ko$\rightarrow$En direction, \hcxt achieves a competitive xCOMET score as 90.3, closely approaching the performance of Qwen3 32B model (92.8). Furthermore, in the opposite direction (En$\rightarrow$Ko), \hcxt surpasses all other models with a score of 85.8. This result indicates that our training pipeline not only preserves English knowledge but also enhances the model's ability to render it into high-quality Korean. These findings, combined with the consistency results, validate that our data curation can deliver bidirectional translation capability without the extensive computational overhead of full-scale multilingual pre-training.

\section{Extensions}
\subsection{Instilling Vision-Language Reasoning in Korean}

The pursuit of sovereign multimodal AI requires not only proficiency in native languages but also robust capabilities for reasoning across modalities. Given that \hcxt was originally developed and optimized for advanced reasoning in text, can it be effectively extended into vision-grounded reasoning through a dedicated \textit{multimodal post-training pipeline}?

In this subsection, we present a separate experimental branch: Starting from the text SFT pipeline (Section~\ref{text-sft}), we incorporate visual modules and multimodal tuning to construct a vision-language model. This enables direct evaluation of complex vision-language reasoning beyond simple transfer from text-only capabilities. For real-world assessment, we use challenging STEM items from the Korean College Scholastic Ability Test (KCSAT). 
As the KCSAT is administered in Korean and reflects rigorous local standards, 
it is suitable as a test of vision-language reasoning ability in Korean.

Each item is presented to the model as an image containing mathematical expressions, tables, diagrams, and scientific text (See Appendix~\ref{appendix:vision-analysis}). The model must first accurately recognize visual content (e.g., text, layout, object and diagram recognition), then perform multi-step logical reasoning. Here, \textit{vision-language reasoning} refers to this integrated process of visual understanding and abstract problem-solving, beyond perceptual recognition alone.

\shortsection{Architecture and Training}
For vision-language reasoning, we augment the LLM backbone with a visual encoder module, similar to the architecture in our previous work, HyperCLOVA X SEED~\citep{hyperclovaxseedvision2024}.
The architecture is composed of:
\begin{itemize}
    \item \textbf{Vision Encoder:} SigLIP-2~\citep{tschannen2025siglip}, operating at 512$\times$512 pixels per grid.
    \item \textbf{Vision-Language Model Architecture:} LLaVA-1.5-HD-based framework~\citep{Liu_2024_CVPR} with C-Abstractor~\citep{cha2023honeybee} connector mechanism, supporting up to 1.57M total pixels distributed over 6 grids.
\end{itemize}

The training pipeline extends our previous protocol~\citep{kim2024hyperclovaxvision} by inserting a dedicated vision SFT stage between text SFT and multimodal RLHF. More concretely, after pre-training on large-scale text corpora, we first apply SFT on instruction-oriented text data, then conduct multimodal SFT with paired image-text data, and finally perform multimodal RLHF on both text-only and multimodal instructions. This change to RLHF---incorporating vision-language samples in addition to text---distinguishes our ablation pipeline from standard text-only approaches. Reasoning Mode (Section~\ref{reasoning-mode}) is toggled via explicit prompting throughout both model training and inference, with ablations performed using both reasoning-enabled and baseline prompt formats.

\shortsection{Experiments}
We evaluate vision-language reasoning performance on the multimodal Korean Educational Test benchmark~\citep{park-kim-2025-evaluating}, with primary focus on its most difficult subset: the KCSAT STEM subjects. This evaluation set comprises 206 items spanning mathematics, physics, chemistry, earth science, and biology, each requiring advanced logical and visual inference at both basic and advanced levels. The KCSAT is internationally recognized for its depth and rigor, making it an exemplary proxy for high-stakes, real-world STEM reasoning.

Our experiments compare \hcxt with Vision against leading contemporary closed APIs in the multimodal LLM space---specifically GPT-4 Turbo with Vision~\citep{openai2024gpt4technicalreport}, GPT-4o~\citep{openai2024gpt4ocard}, GPT-4.1~\citep{openai2024gpt4technicalreport} and OpenAI-o1~\citep{openai2024openaio1card}. All models are assessed under strictly identical protocols using the same visual and textual input, ensuring a fully standardized evaluation environment.

\shortsection{Results and Analysis}
As summarized in Table~\ref{tab:kcast-benchmarks}, \hcxt attains an overall accuracy of 46.4\% on the KCSAT STEM benchmark, outperforming GPT-4.1 (40.3\%) and approaching the performance of GPT-o1 (50.9\%). Disabling Reasoning Mode notably causes performance to drop to 21.7\%, supporting the conclusion that advanced reasoning skills acquired during language pretraining are crucial and can be effectively extended to vision-centric STEM challenges when combined with specialized multimodal tuning. Further qualitative analyses and representative sample outputs are provided in Appendix~\ref{appendix:vision-analysis}.

On the other hand, we note a modest trade-off: adding multimodal SFT introduces a slight decrease in text-only reasoning performance, underscoring the inherent difficulty of jointly optimizing for both modalities. Achieving a balanced sovereign AI---excelling in both unimodal and multimodal reasoning---remains open for further tuning and methodological advances. These observations highlight that effective vision-language reasoning, especially in STEM, demands robust integration of visual parsing and native multi-step logic. As a next step, our future work will extend the model’s capabilities towards unified, native reasoning across text, vision, and audio.

\begin{table}[t!]
    \centering
    \begin{adjustbox}{max width=\textwidth}
    \begin{tabular}{lccccccccccc}
        \toprule
        \multirow{2}{*}{\textbf{Model}} 
            & \multicolumn{2}{c}{\textbf{Math}} 
            & \multicolumn{2}{c}{\textbf{Physics}} 
            & \multicolumn{2}{c}{\textbf{Chemistry}} 
            & \multicolumn{2}{c}{\textbf{Earth Science}} 
            & \multicolumn{2}{c}{\textbf{Biology}} 
            & \multirow{2}{*}{\textbf{Overall}} \\
        \cmidrule(lr){2-3} \cmidrule(lr){4-5} \cmidrule(lr){6-7} \cmidrule(lr){8-9} \cmidrule(lr){10-11}
          & Basic & Adv. 
          & Basic & Adv. 
          & Basic & Adv. 
          & Basic & Adv. 
          & Basic & Adv. 
          & \\
        \midrule
        GPT-4 Turbo with Vision  & 54.5 & 20.8 & 5.0 & 15.0 & 15.0 & 20.0 & 30.0 & 25.0 & 10.0 & 40.0 & 23.8 \\
        GPT-4o              & 68.2 & 50.0 & 15.0 & 20.0 & 25.0 & 25.0 & 40.0 & 30.0 & 15.0 & 25.0 & 32.0 \\
        GPT-4.1              & 68.2 & 66.7 & 20.0 & 30.0 & 40.0 & 30.0 & 30.0 & 40.0 & 35.0 & 35.0 & 40.3 \\
        OpenAI-o1            & 93.2 & 83.3 & 42.5 & 40.0 & 38.8 & 56.3 & 32.5 & 42.5 & 37.5 & 31.3 & 50.9 \\
        \midrule
        \textbf{HyperCLOVA X THINK} with Vision & 68.2 & 68.1 & 33.3 & 28.3 & 41.7 & 58.3 & 28.3 & 38.3 & 43.3 & 50.0 & 46.4 \\
        \:\:\textit{w/o} Reasoning Mode  & 22.7 & 20.8 & 11.7 & 26.7 & 11.7 & 28.3 & 20.0 & 23.3 & 30.0 & 21.7 & 21.7 \\
        \bottomrule
    \end{tabular}
    \end{adjustbox}
\vspace{0.5em}
\caption{Evaluation of native vision-language reasoning ability on the KCSAT STEM multimodal benchmark~\citep{park-kim-2025-evaluating} by subject and level. The benchmark consists of 206 questions covering five scientific subjects (basic/advanced), with 20 to 24 questions per subject and level. KCSAT is widely regarded for its difficulty, emphasis on scientific reasoning, and its reflection of Korea's high-achieving STEM education system.}
    \label{tab:kcast-benchmarks}
\end{table}

\subsection{Lightening through Pruning and Distillation}

As competition in high-performance model development intensifies, training models with tens or hundreds of billions of parameters using trillions of tokens has become the de facto industry standard.
This large-scale training approach entails high costs and long development cycles, while limiting the ability to respond to rapidly changing service environments. Consequently, learning strategies that can efficiently build LLMs with fewer resources have recently gained attention. These approaches not only reduce costs, but also offer practical advantages in terms of timely model development and operation.

One of the leading approaches combines pruning and knowledge distillation. Pruning reduces model size by removing less important parameters, while distillation is a technique that transfers knowledge learned by large models to lightweight models to maintain performance. Combining these two techniques can achieve both model compression and performance preservation simultaneously.

As a real-world example, \href{https://huggingface.co/naver-hyperclovax/HyperCLOVAX-SEED-Text-Instruct-0.5B}{\hcx SEED 0.5B}, recently released on HuggingFace, is the first open-source model in the \hcx series trained using pruning and knowledge distillation. Despite being similar in size to Qwen2.5-0.5B\cite{qwen2025qwen25technicalreport}, it was trained at approximately 39 times lower cost and outperformed competing models in most benchmarks. Notably, it demonstrated significant performance improvements in Korean language benchmarks. This model offers high practical value as it enables high-performance conversational interfaces even in resource-constrained environments such as mobile applications or smart home devices.

Furthermore, the combination of pruning and distillation can be utilized for efficient production of both lightweight and large models. Depending on the structure of the teacher model used for training, the type of data to be transferred, and the learning strategy, models of various sizes and purposes can be produced. This flexibility is expected to improve usability across future generative AI applications.
Currently, a pruned and distilled version of \hcxt is under preparation to be open-sourced.

\section{Conclusion}
In this report, we introduced \hcxtfull, the first reasoning-focused LLM within \hcx family. It is efficiently trained to achieve two primary objectives: advanced reasoning capabilities and the promotion of sovereign AI for Korea.

Its pre-training dataset comprises approximately 6 trillion high-quality tokens spanning Korean, English, and further enhanced by targeted synthetic Korean data. We employ a Peri-LN Transformer scaled with $\mu$P, ensuring stable scalability and cost-efficient training. A three-stage curriculum enables the model to expand its context window to 128k tokens and demonstrate robust long-form chain-of-thought reasoning. Post-training involves supervised fine-tuning and reinforcement learning with verifiable rewards, utilizing a curated data filtering process to address both detailed reasoning and simple answering tasks.

Experiments demonstrate \hcxtfull's competitive performance against other reasoning models on Korea-centric benchmarks such as KMMLU, CSAT, KoBALT-700, HAERAE-1.0, and KoBigBench. Analysis highlights its highly efficient training cost and its ability to preserve robust bilingual consistency. Furthermore, a vision-augmented variant achieved performance comparable to GPT-4.1 on the KCSAT STEM benchmark.

Our report shows that using additional test-time compute to refine model responses is an effective way to push the limits of model capability and improve compute-cost efficiency. As foundational AI technology gains greater potential to enrich people’s lives and shape the future of digital business, we remain committed to improving reasoning scalability and delivering foundation models that are both powerful and affordable, thereby accelerating both domestic and global AI transformation in businesses.

Note that, while we took necessary measures to improve the safety of \hcxtfull as per NAVER AI Ethics guidelines, the harmlessness of the generated text cannot be fully guaranteed. Thus, the responses may contain toxic remarks, exhibit biases or otherwise harmful content. However, we remain dedicated to responsible AI development and deployment.

Lastly, we plan to open-source a pruned and distilled version of \hcxtfull. This initiative aims to benefit academic and industry partners with limited resources, fostering the future development and utilization of sovereign LLMs.

\bibliographystyle{acl_natbib}
\bibliography{anthology,custom}

\newpage
\appendix
\section{Contributors}

\titlespacing*{\subsection}{0pt}{1ex}{0ex}
\textit{Within each role, names are listed in alphabetical order by last name, followed by the first name.}

\begin{multicols}{2}
\subsection*{Core Contributors}

Sanghwan Bae \\
Minseong Choi \\
Hyunsoo Ha \\
Chiheon Ham \\
Donghoon Ham \\
Jaemin Han \\
Jiwoo Hong\textsuperscript{\textdagger} \\
Youngki Hong \\
Jinbae Im \\
Sookyo In \\
Yeguk Jin \\
Chansong Jo \\
Hwiyeol Jo \\
Shinyoung Joo \\
Jingu Kang \\
Donghyeon Ko \\
Taeho Kil \\
Byeongwook Kim \\
Daehee Kim \\
Donghyun Kim \\
Geewook Kim \\
Hanbyul Kim \\
Hyunwoo Kim \\
Jeonghoon Kim \\
Jungwhan Kim \\
Minkyoung Kim \\
Munhyong Kim \\
Seonghyun Kim \\
Sungdong Kim\textsuperscript{\textdagger} \\
Sungju Kim \\
Yoonsik Kim \\
You Jin Kim \\
Donghyun Kwak \\
Beomseok Kwon \\
Bado Lee \\
Byungwook Lee \\
Gichang Lee \\
Hodong Lee \\
Injae Lee \\
Jaehong Lee \\
Jeong Hyun Lee\textsuperscript{\textdagger} \\
Jieun Lee \\
Joosung Lee \\
Min Young Lee \\
Noah Lee \\
Sang-Woo Lee\textsuperscript{\textdagger} \\
Yehbin Lee \\
Yujeong Lee \\
Taehong Min \\
Kiyoon Moon \\
JeongYeon Nam \\
Yeontaek Oh \\
Cheonbok Park \\
Joonsuk Park \\
Kyuyon Park \\
Sanghee Park \\
Ahreum Seo \\
Seunghyun Seo \\
Suk Min Seo \\
Seongjin Shin \\
Ka Yeon Song \\
Nako Sung \\
Moonbin Yim \\
Kang Min Yoo \\
Taehwan Yoo \\
MyungIn You \\
Hangyeol Yu

\subsection*{Contributors}

Sang Min An  \\
Jeongin Bae \\
Chongho Cha  \\
Eungsup Cho  \\
Haesong Cho  \\
Saerim Cho  \\
Hyungwook Choi  \\
Jaepil Choi \textsuperscript{\textdagger} \\
Sanghyuk Choi  \\
Jaehyeok Doo \textsuperscript{\textdagger} \\
Sungbum Hong  \\
Seongchan Hwang  \\
Donghoon Jang  \\
Genie Jang  \\
Junseo Jang  \\
Heewon Jeon  \\
Mina Jeon  \\
Kyeongseok Jeong  \\
Yelim Jeong  \\
Myunggeun Ji  \\
Youngkyun Jin  \\
Ara Jo  \\
Hyunhoon Jung  \\
Kwangsun Jung  \\
Seunghwan Jung  \\
Dain Kim \textsuperscript{\textdagger} \\
Dong Gyun Kim  \\
Eunchul Kim  \\
Ginam Kim  \\
Hyomin Kim  \\
Hyunwook Kim  \\
Jihye Kim  \\
Jiseob Kim \\
Jonghak Kim  \\
Joonghoon Kim \textsuperscript{\textdagger} \\
Minseung Kim  \\
Minyoung Kim  \\
Singon Kim  \\
Soyoon Kim  \\
Taeyong Kim  \\
Yonghee Kim  \\
Youngjun Kim  \\
Ohsung Kwon  \\
Yoo Hwan Kwon  \\
Youngjin Kwon  \\
Dagyeong Lee  \\
Dughyun Lee  \\
Gayoung Lee  \\
Ha Ram Lee  \\
Hagyeong Lee  \\
Jeonghyun Lee  \\
Jonghyun Lee  \\
Jongjin Lee  \\
Joonhyung Lee \\
Junghoon Lee  \\
Seulgi Lee  \\
Soeun Lee  \\
Sujin Lee  \\
Sungwoo Lee  \\
Yesol Lee  \\
Youngbeom Lee  \\
Taemin Lim  \\
Kyeong Min Nam  \\
Biro Oh  \\
Solgil Oh  \\
Gunho Park \\
Wonkyeong Park  \\
Jieun Shin  \\
Wonkwang Shin  \\
Chiyun Song  \\
Hae Jin Song \\
Minchul Song  \\
Jisung Wang \\
Sukwon Yeo  \\
Hwanhee Yoo  \\
Wonjoon You  \\
Uiseon Yu  \\

\end{multicols}

\textsuperscript{\textdagger} Work done while at NAVER Cloud.

\clearpage
\newpage

\section{Performance on Math\&Coding Benchmarks}\label{appendix:math_coding_benchmark}

\begin{table*}[!h]\centering\small
\renewcommand{\arraystretch}{1.6}%
\resizebox{\textwidth}{!}{%
\begin{tabular}{p{2cm}p{2cm}>{\columncolor{customblue}\centering\arraybackslash}p{1.5cm}
>{\centering\arraybackslash}p{1.5cm}
>{\centering\arraybackslash}p{1.5cm}
>{\centering\arraybackslash}p{1.5cm}
>{\centering\arraybackslash}p{1.5cm}}
\toprule
\textbf{Category} & \textbf{Benchmarks} & \textbf{HCX THINK} & \textbf{Qwen3} & \textbf{Qwen3} & \textbf{QwQ} & \textbf{EXAONE Deep} \\
 & & ( - ) & (32B) & (14B) & (32B) & (32B) \\
\midrule
\multirow{4}{*}{\makecell{Math}} & \textbf{GSM8K} & 95.5 & 95.9 & 95.9 & \bf 96.2 & 95.5 \\
 & \textbf{GSM8K-ko} & 92.1 & \bf 93.4 & 92.6 & 92.7 & 91.8 \\
 & \textbf{MATH500} & 95.2 & 97.2 & 96.8 & \bf 98.0 & 96.0 \\
 & \textbf{MATH100-ko} & 90.5 & 92.9 & 92.9 & 92.9 & \bf 93.9 \\
\midrule
\multirow{2}{*}{Coding} & \textbf{HumanEval} & 95.7 & \bf 96.9 & 95.7 & 90.8 & 95.1 \\
 & \textbf{MBPP} & 89.9 & 78.6 & \bf 90.8 & 90.1 & 87.5 \\
\bottomrule
\end{tabular}%
} 
\caption{Performance comparison of language models on math and coding benchmarks. Differences among the baseline reasoning models are marginal, as these benchmarks are approaching saturation.}
\label{tab:math_coding_benchmark}
\end{table*}

\section{KoBigBench Qualitative Examples}
\label{appendix:text-only-analysis}

\begin{tcolorbox}[colback=cyan!1!white, colframe=gray!10!cyan,
  breakable, 
  title=HyperCLOVA X THINK (Translated Model Input\&Output -- English)]
\texttt{<|im\_start|>tool\_list}\\
\texttt{<|im\_end|>}\\
\texttt{<|im\_start|>system}\\
The following are multiple choice questions with explanations and/or answers about social-ecofin201.\texttt{<|im\_end|>}\\
\texttt{<|im\_start|>user}\\
Question: What is the most reasonable measure the government can take to prevent the "transaction cliff" shock in the real estate market?\\
Read the question again: What is the most reasonable measure the government can take to prevent the "transaction cliff" shock in the real estate market?\\
Options:\\
(A). Increase real estate transaction tax rate\\
(B). Increase property tax rate\\
(C). Strengthen safety inspections for reconstruction\\
(D). Lift designation of regulated areas\texttt{<|im\_end|>}\\
\texttt{<|im\_start|>assistant/think}\\
This question asks for the most reasonable measure the government can take to address the "transaction cliff" problem in the real estate market today. We have four options from A to D. I need to examine each option to see which one would help prevent a transaction cliff.\\

First, I need to understand exactly what a "transaction cliff" is. It seems to refer to a phenomenon where real estate transaction volume rapidly decreases. This term is typically used when the real estate market is in a slump or when regulations are tightened, leading to a contraction in transactions. Therefore, to prevent this, the government would likely need to implement policies that stimulate transactions.\\

Let's analyze each option one by one.\\

(A) Increase real estate transaction tax rate: Transaction taxes refer to taxes like acquisition tax or registration tax paid when buying or selling real estate. Increasing the transaction tax rate would increase transaction costs, which would likely further reduce transactions. This would worsen the transaction cliff rather than prevent it. So, A does not seem like an appropriate measure.\\

(B) Increase property tax rate: Property tax refers to taxes paid while owning a home, such as property tax or comprehensive real estate tax. Increasing property taxes might lead some homeowners to put their properties on the market due to the tax burden. However, conversely, existing owners might also choose to hold onto their properties. It's also somewhat ambiguous whether an increase in property taxes would directly stimulate transactions. There's also an argument that increasing property taxes could lead to a decrease in transactions itself. So, B also seems inappropriate for preventing a transaction cliff.\\

(C) Strengthen safety inspections for reconstruction: Strengthening safety inspections for reconstruction would reduce the supply through redevelopment. However, whether this is relevant depends on whether the transaction cliff is due to a supply shortage or other factors. If the lack of transactions is due to insufficient new supply through reconstruction, then relaxation, not strengthening, might be needed. But the question asks for a measure to prevent a transaction cliff, and strengthening safety inspections for reconstruction could further reduce supply, potentially negatively impacting the market. Therefore, C also seems inappropriate.\\

(D) Lift designation of regulated areas: Regulated areas refer to regions where subscription, loan, and resale restrictions apply. Lifting the designation of regulated areas would ease transaction regulations in those areas, which could stimulate transactions, right? For example, if loan regulations are eased or resale restrictions are removed, it would become easier to conduct transactions, potentially increasing transaction volume. Therefore, lifting the designation of regulated areas could help prevent a transaction cliff.\\

However, even with this thought, I need to consider more specifically how lifting the designation of regulated areas actually impacts transaction activation. If regulated areas are lifted, they would be removed from speculative overheated zones or adjustment target areas, leading to relaxed subscription conditions, increased loan limits, and even allowing multiple homeowners to get loans, thereby easing transaction-related regulations. This could increase buying demand and lead to more active transactions. On the other hand, there might be concerns that lifting regulated areas could induce speculation and cause market overheating. However, the question specifically asks for a measure to prevent a transaction cliff, i.e., a rapid decrease in transactions, so regulatory easing seems appropriate.\\

Comparing each option, A is inappropriate because it increases transaction costs and reduces transactions. B might lead to property sales due to increased holding tax but is less directly related to transaction activation. C could lead to reduced supply and further decrease transactions. D could promote transactions by easing regulations, so D seems to be the most reasonable measure.\\

However, one lingering question is whether lifting the designation of regulated areas can be a direct solution to the transaction cliff. For example, if the current real estate market is in a slump and transactions have decreased, easing regulations could stimulate demand and increase transactions. Conversely, if the market was overheated and regulations were tightened, easing them could lead to re-overheating, but the question is about preventing a transaction cliff, a decrease in transactions, so regulatory easing seems correct.\\

Also, another possibility is that if the transaction cliff is due to external factors like rising interest rates or an economic downturn, the government's regulatory easing alone might have limited effect. However, the question asks to choose from government measures, so it's about selecting the most appropriate policy tool.\\

In summary, to increase transactions, lowering transaction costs or easing regulations would be helpful. Increasing the transaction tax rate (A) increases costs and is therefore opposite. Increasing the property tax rate (B) increases the burden of ownership, which might incentivize selling, but it's less directly linked to transaction activation. Strengthening safety inspections for reconstruction (C) could lead to a decrease in supply and thus negatively impact transaction activation. Lifting the designation of regulated areas (D) eases regulations and can promote transactions, so it seems to be the most appropriate measure.\\

However, I need to re-evaluate whether the assumption that lifting the designation of regulated areas helps activate transactions is correct. Regulated areas are designated as speculative overheated or adjustment target areas. Lifting these designations can ease loan regulations, resale restrictions, etc., making transactions easier. For example, if the resale restriction period is shortened, short-term trading becomes possible, which can increase transaction volume. If loan regulations are eased, real demanders can more easily access housing purchases, which can stimulate transactions. Therefore, lifting the designation of regulated areas can help alleviate the transaction cliff.\\

On the other hand, if the current transaction cliff is due to reduced liquidity from interest rate hikes or decreased demand due to an economic downturn, even regulatory easing might not be enough to increase transactions. However, the question asks for a government measure to prevent the transaction cliff in the real estate market, so among the given options, D seems the most appropriate.\\

Let's double-check the other options.\\

(A) Increase real estate transaction tax rate: This would definitely increase transaction costs and reduce transactions, thus deepening the transaction cliff rather than preventing it. So, A is incorrect.\\

(B) Increase property tax rate: While some argue that increasing property taxes might make homeowners feel the tax burden and put properties on the market, conversely, they might also hold onto their properties. Also, increasing property taxes primarily affects multi-homeowners or owners of high-value properties. While their putting properties on the market could increase supply and activate transactions, this can vary depending on the situation. However, generally, an increase in property taxes is more focused on price stabilization through increased supply rather than transaction activation. Therefore, B might not be an appropriate direct measure to solve the transaction cliff itself.\\

(C) Strengthen safety inspections for reconstruction: Expanding supply through reconstruction can contribute to market stability by increasing housing supply. However, strengthening safety inspections makes reconstruction more difficult, thus reducing supply. Reduced supply can lead to upward price pressure but is a separate issue from transaction activation. If the lack of transactions is due to a supply shortage, then supply needs to be increased, but if the current transaction cliff is due to other factors, this measure would be ineffective. Therefore, C seems inappropriate for activating transactions.\\

(D) Lift designation of regulated areas: As I considered earlier, easing regulations can make transactions easier, potentially increasing transaction volume, so D seems the most appropriate.\\

Furthermore, recalling the actual impact of lifting regulated area designations on transaction activation in Korean real estate policy, for example, the government lifted some regulated areas in late 2022 or early 2023. At that time, the real estate market was cooling, and the purpose was to stimulate transactions by easing regulations. Therefore, in actual cases, lifting regulated area designations has been used as a policy to alleviate transaction cliffs, which increases the likelihood that D is the correct answer in this problem.\\

In conclusion, analyzing each option, (D) Lifting designation of regulated areas seems to be the most reasonable measure.
\texttt{<|im\_end|>}
\texttt{<|im\_start|>assistant}\\
Answer: D\texttt{<|im\_end|>}
\end{tcolorbox}

\begin{tcolorbox}[colback=cyan!1!white, colframe=gray!10!cyan,
  breakable, 
  title=HyperCLOVA X THINK (Model Input\&Output -- Korean)]
    \texttt{<|im\_start|>tool\_list}\\
    \texttt{<|im\_end|>}
    \texttt{<|im\_start|>system}\\
The following are multiple choice questions with explanations and/or answers about social-ecofin201.\texttt{<|im\_end|>}\\
    \texttt{<|im\_start|>user}\\
Question: 부동산 시장의 '거래 절벽' 충격을 막기 위해 정부가 취할 수 있는 조치로 가장 합리적인 것은?\\
Read the question again: 부동산 시장의 '거래 절벽' 충격을 막기 위해 정부가 취할 수 있는 조치로 가장 합리적인 것은?\\
Options:\\
 (A). 거래세율 인상\\
 (B). 보유세율 인상\\
 (C). 재건축 안전진단 강화\\
 (D). 규제지역 해제
\texttt{<|im\_end|>}\\
\texttt{<|im\_start|>assistant/think}\\
오늘 부동산 시장의 '거래 절벽' 문제를 해결하기 위해 정부가 취할 수 있는 조치 중에서 가장 합리적인 것을 고르는 문제네요. 문제는 선택지 A부터 D까지 네 가지 옵션이 주어졌어요. 각 옵션을 하나씩 살펴보면서 어떤 것이 거래 절벽을 막는 데 도움이 될지 생각해봐야겠어요.\\

먼저, '거래 절벽'이 무엇인지부터 정확히 이해해야 할 것 같아요. 거래 절벽이란 부동산 거래량이 급격히 감소하는 현상을 말하는 것 같은데, 보통 부동산 시장이 침체기에 빠지거나 규제가 강화되어 거래가 위축될 때 사용되는 용어인 것 같아요. 그렇다면 정부가 이를 막기 위해서는 거래를 활성화시킬 수 있는 정책을 펴야 하지 않을까 싶네요.\\

각 선택지를 하나씩 분석해 볼게요.\\

(A) 거래세율 인상: 거래세란 부동산을 사고 팔 때 내는 취득세나 등록세 같은 것을 말하는 것 같아요. 거래세율을 인상하면 거래 비용이 늘어나게 되므로, 거래를 더 줄이는 효과가 있을 것 같은데. 그러면 거래 절벽을 막기보다는 오히려 악화시키지 않을까? 따라서 A는 적절한 조치가 아닐 것 같아요.\\

(B) 보유세율 인상: 보유세는 주택을 소유하고 있는 동안 내는 세금, 예를 들어 재산세나 종합부동산세 같은 걸 말하죠. 보유세를 올리면 집을 가진 사람들이 세금 부담을 느껴 매물을 내놓을 수도 있겠지만, 반대로 생각하면 기존 소유자들이 매물을 거둬들이게 될 수도 있을 것 같아요. 하지만 보유세 인상이 거래 활성화에 직접적인 영향을 미칠지는 좀 애매한 것 같아요. 특히 보유세를 올리면 거래 자체가 줄어들 수 있다는 의견도 있을 것 같고. 그래서 B도 거래 절벽을 막는 데는 부적절할 것 같네요.\\

(C) 재건축 안전진단 강화: 재건축 안전진단을 강화하면 재건축을 통한 공급이 줄어들게 되겠죠. 그런데 거래 절벽이 공급 부족으로 인한 것인지, 아니면 다른 요인 때문인지에 따라 다를 것 같아요. 만약 재건축을 통한 신규 공급이 부족해서 거래가 안 된다면 강화가 아니라 완화가 필요할 수도 있지 않을까요? 하지만 문제에서는 거래 절벽을 막기 위한 조치라고 했으니, 재건축 안전진단 강화는 공급을 더 줄여서 오히려 시장에 부정적인 영향을 줄 수도 있을 것 같아요. 따라서 C도 적절하지 않을 것 같네요.\\

(D) 규제지역 해제: 규제지역이란 청약, 대출, 전매 제한 등이 적용되는 지역을 말하는 것 같아요. 규제지역을 해제하면 해당 지역에서의 거래 규제가 완화되어 거래가 활성화될 수 있지 않을까? 예를 들어, 대출 규제가 완화되거나 전매 제한이 없어지면 거래를 하기 쉬워져서 거래량이 늘어날 수 있을 것 같아요. 따라서 규제지역 해제는 거래 절벽을 막는 데 도움이 될 수 있을 것 같네요.\\

하지만 이렇게 생각해보지만, 실제로 규제지역 해제가 거래 활성화에 어떤 영향을 미치는지 좀 더 구체적으로 알아볼 필요가 있을 것 같아요. 규제지역이 해제되면 투기과열지구, 조정대상지역 등에서 벗어나게 되어서 청약 조건이 완화되고, 대출 한도가 늘어나거나, 다주택자도 대출을 받을 수 있게 되는 등 거래 관련 규제가 완화되겠죠. 이렇게 되면 매수 수요가 늘어나 거래가 활발해질 수 있을 것 같아요. 반면에 규제지역 해제가 오히려 투기를 유발하여 시장 과열을 초래할 수도 있다는 우려도 있을 수 있지만, 문제에서는 거래 절벽, 즉 거래가 급격히 줄어드는 현상을 막기 위한 조치이므로 규제 완화가 도움이 될 것 같아요.\\

그래서 각 선택지를 비교해보면, A는 거래 비용을 늘려 거래를 줄이므로 부적절하고, B는 보유세 인상으로 매물 출회가 있을 수도 있지만 직접적인 거래 활성화보다는 다른 효과를 가져올 수 있으며, C는 공급 감소로 거래를 더 줄일 수 있고, D는 규제 완화로 거래를 촉진할 수 있으므로 가장 합리적인 것은 D가 아닐까 생각해요.\\

하지만 여기서 한 가지 의문이 드는 점이 있어요. 규제지역 해제가 거래 절벽을 해결하는 직접적인 방법이 될 수 있을까요? 예를 들어, 현재 부동산 시장이 침체기라서 거래가 줄어든 상황이라면, 규제를 풀어주는 것이 수요를 자극하여 거래를 늘릴 수 있을 거예요. 반면에 시장이 과열되어서 규제를 강화한 상황이라면, 규제를 완화하면 다시 과열될 수 있지만, 문제에서는 거래 절벽, 즉 거래 감소 상황을 막기 위한 것이므로 규제 완화가 맞을 것 같아요.\\

또한 다른 가능성으로는, 거래 절벽이 금리 상승이나 경기 침체 등 외부 요인에 의한 것이라면 정부의 규제 완화만으로는 효과가 제한적일 수 있지만, 문제에서는 정부의 정책 중에서 선택하라고 했으니 정책적 수단 중에서 가장 적절한 것을 고르는 문제일 거예요.\\

정리하자면, 거래를 늘리기 위해서는 거래 비용을 낮추거나 규제를 완화하는 것이 도움이 될 텐데, 거래세율 인상(A)은 비용을 높이므로 반대이고, 보유세 인상(B)은 보유 부담을 높여 매도 유인은 있지만 거래 활성화와는 직접 연관이 적을 것 같아요. 재건축 안전진단 강화(C)는 공급 감소로 작용할 수 있으니 역시 거래 활성화에 부정적일 것 같고, 규제지역 해제(D)는 규제를 완화하여 거래를 촉진할 수 있으므로 가장 적절한 조치로 보여요.\\

하지만 여기서 규제지역 해제가 거래 활성화에 도움이 된다는 가정이 맞는지 다시 한 번 생각해봐야 할 것 같아요. 규제지역으로 지정된 지역은 투기과열이나 조정대상지역 등인데, 이 지역들을 해제하면 대출 규제, 전매 제한 등이 완화되어 거래가 용이해질 수 있습니다. 예를 들어, 전매 제한 기간이 줄어들면 단기 매매가 가능해져서 거래량이 늘어날 수 있고, 대출 규제가 완화되면 실수요자들이 주택 구입에 더 쉽게 접근할 수 있게 되어 거래가 활성화될 수 있겠죠. 따라서 규제지역 해제는 거래 절벽을 완화하는 데 도움이 될 수 있을 것 같아요.\\

반면에, 만약 현재 거래 절벽이 금리 인상으로 인한 유동성 축소나 경기 침체에 따른 수요 감소 때문이라면, 규제 완화로도 거래를 늘리기 어려울 수 있지만, 문제에서는 부동산 시장의 거래 절벽을 막기 위한 정부의 조치 중에서 선택하라고 했으므로, 주어진 선택지 중에서는 D가 가장 적절하다고 봐야 할 것 같아요.\\

그럼 다른 선택지들을 다시 한 번 체크해보죠.\\

(A) 거래세율 인상: 이건 확실히 거래 비용을 증가시켜 거래를 줄이므로, 거래 절벽을 막기보다는 심화시킬 것 같아요. 따라서 A는 틀린 답.\\

(B) 보유세율 인상: 보유세를 올리면 집주인들이 세금 부담을 느껴 매물을 내놓을 수 있다는 의견도 있지만, 반대로 매물을 내놓지 않고 버티기를 할 수도 있어요. 또한, 보유세 인상은 주로 다주택자나 고가 주택 소유자에게 타격을 줄 텐데, 이들이 매물을 내놓으면 공급이 늘어나 거래가 활성화될 수도 있겠지만, 이는 상황에 따라 다를 수 있어요. 하지만 일반적으로 보유세 인상은 거래 활성화보다는 공급 증가를 통한 가격 안정화에 더 초점이 맞춰져 있을 것 같아요. 따라서 거래 절벽 자체를 해결하기 위한 직접적인 조치로는 B가 적절하지 않을 수 있어요.\\

(C) 재건축 안전진단 강화: 재건축을 통한 공급 확대는 주택 공급을 늘려서 시장 안정에 기여할 수 있지만, 안전진단을 강화하면 재건축이 어려워져서 공급이 줄어들게 되죠. 공급 감소는 가격 상승 압력으로 작용할 수 있지만, 거래 활성화와는 별개 문제일 수 있어요. 만약 공급이 부족해서 거래가 안 된다면 공급을 늘려야 하지만, 현재 거래 절벽은 공급 부족보다는 다른 요인 때문이라면 이 조치는 효과가 없을 거예요. 따라서 C는 거래 활성화를 위한 조치로는 부적절해 보여요.\\

(D) 규제지역 해제: 앞서 생각한 것처럼 규제를 완화하면 거래가 용이해져서 거래량이 늘어날 수 있으므로 D가 가장 적절하다고 생각됩니다.\\

하지만 실제로 한국의 부동산 정책에서 규제지역 해제가 거래 활성화에 미친 영향은 어땠는지 기억을 더듬어 보면, 예를 들어 2022년 말이나 2023년 초에 정부가 일부 규제지역을 해제한 적이 있었어요. 그 당시에는 부동산 시장이 냉각되어 있었기 때문에 규제를 완화해서 거래를 촉진하려는 목적이었을 거예요. 따라서 실제 사례에서도 규제지역 해제가 거래 절벽 완화에 도움이 되는 정책으로 사용된 바 있으니, 이 문제에서도 D가 정답일 가능성이 높아 보여요.\\

결론적으로, 각 옵션을 분석해보면 (D) 규제지역 해제가 가장 합리적인 조치로 보입니다.\texttt{<|im\_end|>}
\texttt{<|im\_start|>assistant}
Answer:	D\texttt{<|im\_end|>}
\end{tcolorbox}

\begin{tcolorbox}[colback=cyan!1!white, colframe=gray!10!cyan,
  breakable, 
  title=HyperCLOVA X THINK (Translated Model Input\&Output -- English)]
  \texttt{<|im\_start|>tool\_list}\\
\texttt{<|im\_end|>}
\texttt{<|im\_start|>system}\\
The following are multiple choice questions with explanations and/or answers about commonsense-context30.\texttt{<|im\_end|>}\\
\texttt{<|im\_start|>user}\\
Question: Context: Everyone eventually came back to me and made me feel loved. It was truly wonderful. It’s amazing and touching that I have such wonderful friends by my side. I think it felt even more special because I hadn't seen them in such a long time. It's strangely surprising to have returned like this, but meeting people again is really a great thing. Especially Donghwan.\\
Question: If I hadn't left, how would I have felt?\\
Read the question again: Context: Everyone eventually came back to me and made me feel loved. It was truly wonderful. It’s amazing and touching that I have such wonderful friends by my side. I think it felt even more special because I hadn't seen them in such a long time. It's strangely surprising to have returned like this, but meeting people again is really a great thing. Especially Donghwan.\\
Question: If I hadn't left, how would I have felt?\\
Options:\\
(A). Donghwan would have hated me, and nothing would have happened.\\
(B). I wouldn't have missed these people this much.\\
(C). It's still joyful and touching to have friends who hurt me.\\
(D). It would have felt strange not to have left home at all.\texttt{<|im\_end|>}\\
\texttt{<|im\_start|>assistant/think}\\
Today, the user provided context and a question, requiring me to select the most appropriate answer. I'll first carefully examine the context provided. The speaker mentions feeling loved and touched because friends returned to their side, emphasizing how special it felt due to the long absence, especially highlighting Donghwan. The speaker mentions that returning felt strangely surprising but meeting people again was truly great.\\

The question asks, "If I hadn't left, how would I have felt?" Currently, the speaker is moved because of the reunion after leaving and returning. If they had never left, the intense feelings related to the reunion might not have existed.\\

Analyzing each option individually:\\

(A). Donghwan would have hated me, and nothing would have happened.\\
This doesn't align with the context. There is no indication in the context that Donghwan or others would have felt negatively if the speaker hadn't left.\\

(B). I wouldn't have missed these people this much.\\
This is logical. The speaker emphasizes that the long absence made the reunion more special and touching. Without leaving, the feeling of longing would naturally be less intense.\\

(C). It's still joyful and touching to have friends who hurt me.\\
This is incorrect. The context never indicates that the friends were hurtful. Rather, it focuses on positive and warm feelings upon reunion.\\

(D). It would have felt strange not to have left home at all.\\
The context describes the return as "strangely surprising," suggesting that leaving and returning was unusual. If the speaker had never left, it might have felt normal rather than strange to remain home. Therefore, option D is somewhat ambiguous but doesn't strongly align with the context.\\

Given the context and the reasoning above, option (B) stands out as the most appropriate choice. The emotional impact of missing friends after a long absence is clearly highlighted, and thus, had the speaker never left, these strong feelings of longing would not have existed.\\

Hence, the correct answer is option (B).\texttt{<|im\_end|>}\\
    \texttt{<|im\_start|>assistant}\\
    Answer:	B\texttt{<|im\_end|>}
\end{tcolorbox}

\begin{tcolorbox}[colback=cyan!1!white, colframe=gray!10!cyan,
  breakable, 
  title=HyperCLOVA X THINK (Model Input\&Output -- Korean)]
    \texttt{<|im\_start|>tool\_list}\\
\texttt{<|im\_end|>}
\texttt{<|im\_start|>system}\\
The following are multiple choice questions with explanations and/or answers about commonsense-context30.\texttt{<|im\_end|>}\\
\texttt{<|im\_start|>user}\\
Question: 맥락: 모두가 결국 저한테 돌아와줬고 제가 사랑받고 있다고 느끼게 했어요. 진짜 최고였네요. 진짜 이런 멋진 친구들이 제 곁에 있다는게 너무 좋고 감동적이에요. 특히 오랫동안 못봐서 더 그랬던거 같아요. 이렇게 돌아오게 된게 이상할 정도지만 사람들이랑 다시 만난다는 것은 진짜 좋은 일이에요. 특히 동환이요.\\
질문: 만약 내가 떠나지 않았더라면 내가 어떻게 느꼈을까?\\
Read the question again: 맥락: 모두가 결국 저한테 돌아와줬고 제가 사랑받고 있다고 느끼게 했어요. 진짜 최고였네요. 진짜 이런 멋진 친구들이 제 곁에 있다는게 너무 좋고 감동적이에요. 특히 오랫동안 못봐서 더 그랬던거 같아요. 이렇게 돌아오게 된게 이상할 정도지만 사람들이랑 다시 만난다는 것은 진짜 좋은 일이에요. 특히 동환이요.\\
질문: 만약 내가 떠나지 않았더라면 내가 어떻게 느꼈을까?\\
Options:\\
 (A). 동환이가 나를 미워했을거라 아무 일도 없었을 것이다.\\
 (B). 이만큼 이 사람들을 그리워하진 않았을 것이다.\\
 (C). 상처 주는 친구들이 있다는 것이 여전히 기쁘고 감동적이다.\\
 (D). 집을 전혀 떠나지 않았다는 것이 이상하게 느껴졌을 것이다.\texttt{<|im\_end|>}\\
\texttt{<|im\_start|>assistant/think}\\
오늘 사용자가 맥락과 질문을 주고, 그에 맞는 답을 선택하는 문제를 내주었어. 먼저 맥락을 자세히 읽어봐야겠지. 맥락을 보면, 주인공은 친구들이 돌아와주고 사랑받는다고 느껴서 감동적이고 좋았다고 말하고 있어. 특히 오랫동안 못 봤던 친구들, 특히 동환이를 다시 만나게 되어서 더 그랬다고 해. 그리고 "이렇게 돌아오게 된게 이상할 정도지만"이라고 했어. 이게 중요한 부분일 것 같아.\\

질문은 "만약 내가 떠나지 않았더라면 내가 어떻게 느꼈을까?"야. 현재 상황은 주인공이 떠났다가 다시 돌아왔고, 그로 인해 친구들이 돌아와줘서 감동받은 상황이야. 그런데 만약 떠나지 않았다면 어떻게 느꼈을까를 묻는 거지.\\

선택지를 하나씩 분석해봐야겠어.\\

(A) 동환이가 나를 미워했을거라 아무 일도 없었을 것이다.\\
이건 맥락과 좀 안 맞는 것 같아. 현재 상황은 친구들이 돌아와줘서 좋은 상황인데, 떠나지 않았다면 동환이가 미워했을 거라니? 맥락에서는 떠나서 다시 돌아왔을 때 친구들이 돌아와줬으니, 떠나지 않았다면 친구들이 떠나지 않았을까? 그런데 이건 좀 헷갈리네. 아니면 주인공이 떠났기 때문에 친구들이 돌아온 것일까? 맥락에서 "모두가 결국 저한테 돌아와줬고"라고 했으니까, 주인공이 떠난 후 친구들이 돌아온 걸 수도 있겠지. 그런데 만약 떠나지 않았다면 친구들이 돌아올 필요가 없었을 거고, 그래서 그 감동을 느끼지 못했을 수도 있겠다. 그런데 (A)는 동환이가 미워했을 거라고 하니 맥락과 연결이 잘 안 돼 보여.\\

(B) 이만큼 이 사람들을 그리워하진 않았을 것이다. 
이건 가능할 것 같아. 주인공이 떠났기 때문에 친구들을 오랫동안 못 봤고, 그래서 다시 만났을 때 더 감동적이었던 거잖아. 만약 떠나지 않았다면 오랫동안 못 보지 않았을 테니 그렇게까지 그리워하지 않았을 거라는 말이 되겠지. 맥락에서 "특히 오랫동안 못봐서 더 그랬던거 같아요"라고 했으니, 떠나지 않았다면 그 부분이 없었을 테니까. 이 선택지는 타당해 보여.\\

(C) 상처 주는 친구들이 있다는 것이 여전히 기쁘고 감동적이다.\\
이건 맥락과 맞지 않아. 현재 상황은 친구들이 돌아와서 감동적이지만, 상처 주는 친구들이라면? 맥락에서는 친구들이 돌아와서 좋은 상황이니까, 상처 주는 친구들이 있다는 게 아니라 오히려 돌아와서 좋은 상황이잖아. 따라서 (C)는 부적절한 것 같아.\\

(D) 집을 전혀 떠나지 않았다는 것이 이상하게 느껴졌을 것이다.\\
이건 좀 애매해. 현재 상황에서 "이렇게 돌아오게 된게 이상할 정도지만 사람들이랑 다시 만난다는 것은 진짜 좋은 일이에요"라고 했으니까, 돌아오는 게 이상했지만 좋은 일이라고 했어. 만약 집을 떠나지 않았다면, 돌아오는 상황 자체가 없으니까, "집을 전혀 떠나지 않았다는 것이 이상하게 느껴졌을 것이다"라고 하면, 현재 상황에서 집을 떠난 것이 이상했지만 좋은 거라면, 떠나지 않았을 때는 그 이상함이 없으니 이상하게 느껴지지 않을까? 그런데 이건 좀 추측이 필요해. 맥락에서 "이렇게 돌아오게 된게 이상할 정도"라고 했으니까, 떠나지 않았다면 돌아올 필요가 없으니 그 이상함은 없었을 거고, 따라서 집을 떠나지 않았다는 것이 이상하게 느껴지지 않을 것 같은데. 아니면 원래 집을 떠나는 게 일반적인 상황이라면, 떠나지 않았을 때 오히려 이상하게 느껴질 수도 있을까? 이건 좀 애매한 것 같아.\\

다시 질문으로 돌아가서, "만약 내가 떠나지 않았더라면 내가 어떻게 느꼈을까?"라고 했을 때, 현재 상황은 떠났다가 돌아와서 친구들을 다시 만나 감동받았어. 따라서 떠나지 않았다면 친구들과 떨어져 있지 않았을 테니, 다시 만나는 감동은 없었을 거야. 특히 오랫동안 못 봤기 때문에 더 감동적이었던 부분이 있으므로, 떠나지 않았다면 그만큼 그리워하지 않았을 것(B)이라는 말이 맞을 것 같아.\\

(A)는 동환이가 미워했을 거라니 맥락과 안 맞고, (C)는 상처 주는 친구들이라는 게 말이 안 되고, (D)는 집을 떠나지 않은 것이 이상하게 느껴진다는 건데, 현재 상황에서 집을 떠난 것이 이상했지만 좋은 거라면, 떠나지 않았을 때는 그 이상함이 없으니 이상하지 않을 것 같은데. 따라서 (D)는 애매하고, (B)가 가장 자연스러워 보여.\\

하지만 맥락에서 "이렇게 돌아오게 된게 이상할 정도지만 사람들이랑 다시 만난다는 것은 진짜 좋은 일이에요"라고 했으니, 돌아오는 게 예상 밖의 일이었지만 좋은 일이었다는 거지. 만약 떠나지 않았다면 돌아올 필요가 없으니, 돌아올 일이 없어서 그 좋은 일을 경험하지 못했을 거야. 따라서 "이만큼 이 사람들을 그리워하진 않았을 것이다"라는 (B)가 맞을 것 같아. 왜냐하면 떠나서 오랫동안 못 봤기 때문에 그리워했고, 다시 만나서 감동적이었으니까, 떠나지 않았다면 그렇게까지 그리워하지 않았을 테니까.\\

그래서 정답은 (B)인 것 같아.\texttt{<|im\_end|>}\\
\texttt{<|im\_start|>assistant}\\
Answer:	B\texttt{<|im\_end|>}
\end{tcolorbox}

\section{Qualitative Examples in Vision Adaptation Ablation}
\label{appendix:vision-analysis}

We present below a representative example from the KCSAT STEM vision benchmark.
\textbf{All problems are rendered and presented to the model strictly as images}, without any accompanying text input. This setup requires the model to make use of its full vision capability--not only to recognize and interpret textual content within the images, but also to read and reason about mathematical expressions, tables, and diagrams where present.
For all examples, the user query is standardized as:
\textbf{``이미지를 보고 문제를 풀어주세요.''} (“Please solve the problem by looking at the image.”)

\subsection{KCSAT Biology Basic Example}

\begin{figure}[ht!]
    \centering
    \includegraphics[width=0.6\linewidth]{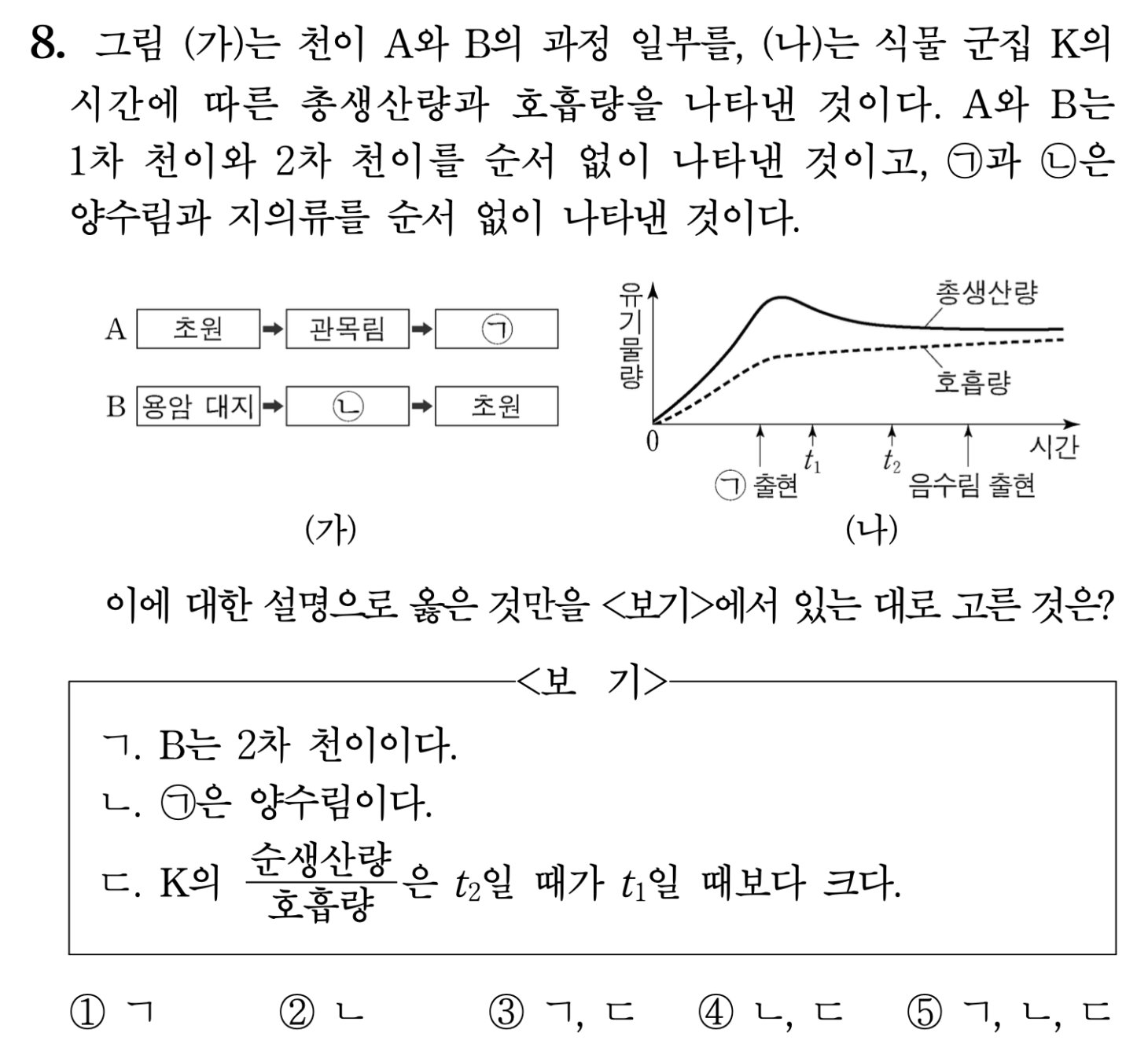}
    \caption{Original KCSAT Sample (in Korean).}
    \label{fig:kcsat-biology-basic-png}
\end{figure}

We present our model’s output on a sample KCSAT Biology Basic problem. Figure~\ref{fig:kcsat-biology-basic-png} shows the original problem image, which is directly provided to the model for analysis and reasoning. The model is required to interpret, reason through, and solve the problem solely based on the image. For reference, we include the model’s original answer, as well as an English translation of the problem for readers. A literal English translation of the model’s output is also provided for comparison.


\begin{tcolorbox}[colback=gray!3!white, colframe=gray!80!black, boxrule=0.3mm, width=0.97\linewidth]
\textbf{Sample Question 8 (Translated)}

The diagrams below represent parts of processes A and B in succession, and a graph depicting the changes in total productivity and respiration amounts of plant community K over time. A and B refer to primary and secondary successions in no particular order, and labels ㄱ and ㄴ refer to hydrosere and xerosere in no particular order.

\medskip

<Process Diagram>    <Graph>

\medskip
\noindent
Which of the following statements are correct based on the above?

\medskip
\noindent
\textbf{(Options)}
\medskip

\fbox{%
  \parbox{0.93\linewidth}{%
    (a) B is a secondary succession.\\
    (b) ㄱ is a hydrosere.\\
    (c) For K, the $\dfrac{\text{total productivity}}{\text{respiration}}$ is greater at $t_2$ than at $t_1$.
  }%
}

\medskip

Select the correct combination(s):

\begin{center}
\begin{tabular}{cccccc}
(1) a         & (2) b         & (3) a, c    & (4) b, c    & (5) a, b, c \\
\end{tabular}
\end{center}

\end{tcolorbox}

\begin{tcolorbox}[colback=cyan!1!white, colframe=gray!10!cyan,
  breakable, 
  title=HyperCLOVA X THINK (Translated Model Output -- English)]
\texttt{<|im\_start|>assistant/think}

Okay, let’s solve this problem. First, we need to look carefully at the figure given in the problem.

First, we need to examine figures (a) and (b) of question 8. According to the problem, A and B are one of primary succession and secondary succession, and ㉠ and ㉡ represent deciduous forest and lichen respectively, in no particular order. Also, it shows a graph of the changes in total productivity and respiration amount of plant community K over time.

Let’s look at (a) first. A is said to go from grassland to shrub forest to ㉠, and B is said to start from lava field, go through ㉡, and reach grassland. Here, we need to figure out which succession process A and B each represent. Also, we need to identify what ㉠ and ㉡ are.

If we think about the succession process, primary succession starts in lifeless areas, and secondary succession is the process of recovery after disturbance in an area where plants already existed. For example, starting on bare land created by disturbances like forest fires or logging is secondary succession. On the other hand, primary succession starts in places like rocks or lava fields, where there isn’t even soil.

In the problem, since A starts from grassland (a place where plants already exist), this would be the secondary succession process. That’s because grassland is an environment with soil and some living organisms. In contrast, B starts from a lava field, which is formed by volcanic activity and likely has no soil. Thus, B would correspond to primary succession. Therefore, the conclusion is that A is secondary succession and B is primary succession... but wait, in the problem it states that A and B represent “primary and secondary in no order,” so I might have thought in reverse, so confirmation is necessary.

Looking again, A: grassland → shrub forest → ㉠

B: lava field → ㉡ → grassland

A already begins with grassland, so it’s a secondary succession process, and B starts from lava field, so it is a primary succession process. So A is secondary, B is primary. Then in the choices, statement ㄱ says “B is secondary succession,” so that’s a wrong statement. So ㄱ is incorrect.

Next, let’s look at what ㉠ and ㉡ are. A process is secondary succession, and generally, the final stage of secondary succession could be a climax, such as a shade-tolerant forest, but depending on the case, it could be something else. However, usually, secondary succession develops into a more complex forest structure over time. But here, A passes through shrub forest and goes to ㉠, so if ㉠ is a deciduous forest, the next stage would be a shade-tolerant forest, but does the problem refer to the final stage or an intermediate one?

On the other hand, B is primary succession, starting from a lava field, going through ㉡, and then to grassland. In the early stage of primary succession, pioneer species such as lichen or moss appear, and then herbaceous plants grow, followed by shrub forest, and eventually deciduous, mixed, and shade-tolerant forests in order. However, since B goes from lava field through ㉡ to grassland, it’s likely that ㉡ is lichen. That’s because in harsh environments like lava fields, what settles first tends to be things like lichen. Then ㉡ is lichen, and ㉠ would be deciduous forest. That’s because, as A undergoes secondary succession, after going through shrub forest it would further develop into deciduous forest. Therefore, ㉠ is deciduous forest and ㉡ is lichen. Accordingly, statement ㄴ, “㉠ is deciduous forest,” is a correct statement.

Next, let’s look at figure (b). It shows changes in total productivity and respiration of plant community K depending on time (t). At time t0, it appears, and up to t1, total productivity increases, then afterwards decreases, while respiration continues to increase, and from t2 onward, it is marked as the appearance of a shade-tolerant forest. That is, before t1, total productivity is greater than respiration, so net productivity has a positive value, after t1, total productivity decreases and becomes similar to or even less than respiration, so net productivity reduces.

Net productivity is total productivity minus respiration, so at t1, total productivity is at its maximum and respiration is relatively low, so net productivity will be very high. But at t2, total productivity is lower and respiration continues to increase, so net productivity will be far lower than at t1. Therefore, the ratio of K’s net productivity to respiration is greater at t1 than at t2. But the choice ㄷ says, “K’s net productivity/respiration is greater at t2 than at t1,” so this is a wrong explanation. So ㄷ is wrong too.

In summary:

- ㄱ: “B is secondary succession.” → B is primary succession, so wrong.

- ㄴ: “㉠ is deciduous forest.” → Correct.

- ㄷ: net productivity/respiration is t2 > t1 → Wrong. t1 is greater.

Therefore, the answer is only ㄴ is correct, so we need to check if there is an option with only ㄴ. The options provided are as follows:

① ㄱ

② ㄴ

③ ㄱ, ㄷ

④ ㄴ, ㄷ

⑤ ㄱ, ㄴ, ㄷ

In this, the case where only ㄴ is correct is option ②, but let’s check again whether there was any calculation mistake.

At t1, total productivity is high and respiration is still low, so net productivity is abundant, and at t2, total productivity has decreased and respiration has increased, so net productivity is lower. Therefore, the ratio of net productivity to respiration, that is, net productivity/respiration, is much greater at t1. For example, if at t1, total productivity is 10 and respiration is 3, net productivity is 7 and the ratio is 7/3 ≈ 2.33. If at t2, total productivity is 5 and respiration is 4, net productivity is 1 and the ratio is 1/4 = 0.25. So t1 is definitely greater. Thus, ㄷ is wrong, and the answer is option ②.

**Final Answer**:
\boxed{(2)}
\texttt{<|im\_end|>}

\texttt{<|im\_start|>assistant}

Analyze figures (a) and (b) to solve the problem.

1. **Analysis of succession process**:

   - **A**: Starts from grassland, passes through shrub forest, and proceeds to ㉠. Since it starts from grassland where plants already exist, it is **secondary succession**.
   
   - **B**: Starts from lava field, passes through ㉡, and proceeds to grassland. Since it starts from a lava field with no soil, it is **primary succession**.
   
   - Therefore, ㄱ ("B is secondary succession.") is an **incorrect** explanation.

2. **Identification of ㉠ and ㉡**:

   - **A** (secondary succession): Passes through shrub forest and proceeds to ㉠, and generally, the late stage of secondary succession is a **deciduous forest**.
   
   - **B** (primary succession): Starts from lava field, passes through ㉡, and proceeds to grassland, with the initial stage being **lichen**.
   
   - Therefore, ㄴ ("㉠ is a deciduous forest.") is a **correct** explanation.

3. **Graph analysis**:

   - **Total productivity** increases until \( t_1 \) and then decreases, while **respiration** continues to increase.
   
   - **Net productivity** is at its highest at \( t_1 \) when total productivity is maximal and respiration is relatively low, so the ratio is high. At \( t_2 \), total productivity decreases and respiration increases, so the net productivity ratio becomes lower.
   
   - Therefore, ㄷ ("\( K \)'s \( \frac{\text{net productivity}}{\text{respiration}} \) is greater at \( t_2 \) than at \( t_1 \)") is an **incorrect** explanation.

In conclusion, the only correct explanation is **ㄴ**.

\boxed{(2)}
\texttt{<|im\_end|>}
\end{tcolorbox}

\begin{tcolorbox}[colback=cyan!1!white, colframe=gray!10!cyan,
  breakable, 
  title=HyperCLOVA X THINK (Original Model Output -- Korean)]
\texttt{<|im\_start|>assistant/think}

좋아, 이 문제를 풀어보자. 일단 문제에서 주어진 그림을 잘 봐야겠지. 

먼저 8번 문제의 그림 (가)와 (나)를 살펴봐야 해. 문제에 따르면 A와 B는 1차 천이와 2차 천이 중 하나이고, ㉠과 ㉡은 양수림과 지의류를 순서 없이 나타낸 거야. 그리고 식물 군집 K의 시간에 따른 총생산량과 호흡량을 그래프로 보여주고 있어.

일단 (가)부터 보자. A는 초원에서 관목림을 거쳐 ㉠으로 가고, B는 용암 대지에서 시작해서 ㉡을 거쳐 초원으로 간다고 되어 있네. 여기서 A와 B가 각각 어떤 천이 과정인지 알아내야 해. 또 ㉠과 ㉡이 무엇인지도 파악해야 하고.

천이 과정을 생각해보면, 1차 천이는 생명이 없는 지역에서 시작하는 거고, 2차 천이는 기존에 식물이 있던 곳에서 교란이 일어난 후에 다시 회복되는 과정이야. 예를 들어 산불이나 벌목 같은 것들로 인해 생긴 빈 땅에서 시작하는 게 2차 천이지. 반면 1차 천이는 바위나 용암 대지처럼 토양조차 없는 곳에서 시작돼.

문제에서 A는 초원(이미 식물이 있는 곳)에서 시작하니까 이건 2차 천이 과정이겠지. 왜냐하면 초원은 이미 토양이 있고 일부 생물이 존재하는 환경이니까. 반면 B는 용암 대지에서 시작하는데, 용암 대지는 화산 활동으로 인해 형성된 곳으로 토양이 없을 가능성이 높아. 그래서 B는 1차 천이 과정에 해당할 거야. 따라서 A는 2차 천이, B는 1차 천이라는 결론이 나오지만... 잠깐, 문제에서는 A와 B가 1차와 2차를 순서없이 나타냈다고 했으니까 내가 반대로 생각했을 수도 있으니 확인 필요.

다시 보면, A: 초원 → 관목림 → ㉠

B: 용암대지 → ㉡ → 초원

A는 이미 초원이 있으므로 2차 천이 과정이고, B는 용암대지에서 시작하므로 1차 천이 과정이다. 그러면 A는 2차, B는 1차가 맞아. 그럼 보기에서 ㄱ은 'B는 2차 천이이다.'라고 했으니 틀린 설명이 되겠지. 그러니까 ㄱ은 틀렸어.

다음으로 ㉠과 ㉡이 뭔지 보자. A 과정은 2차 천이이므로, 일반적으로 2차 천이의 종착 단계는 극상인 음수림이 될 수 있지만, 경우에 따라서는 다른 형태일 수도 있겠지. 하지만 보통 2차 천이는 시간이 지나면 더 복잡한 구조의 숲으로 발달해. 그런데 여기서는 A가 관목림을 거쳐 ㉠으로 가는데, 만약 ㉠이 양수림이라면 그 다음 단계는 음수림일 텐데, 문제에서는 최종 단계를 나타내는 걸까? 아니면 중간 단계일까?

반면 B는 1차 천이로, 용암대지에서 시작하여 ㉡을 거쳐 초원으로 간다. 1차 천이의 초기에는 지의류나 선태식물 같은 개척자들이 나타나고, 그 다음에 초본 식물들이 자라면서 관목림이 되고, 결국에는 양수림, 혼합림, 음수림 순으로 진행되지. 그런데 B는 용암대지에서 ㉡을 거쳐 초원으로 가므로, ㉡은 아마도 지의류일 가능성이 높아. 왜냐하면 용암대지와 같은 척박한 환경에서 가장 먼저 정착하는 것은 지의류 같은 것이기 때문이야. 그렇다면 ㉡은 지의류이고, ㉠은 양수림이 되는 거겠지. 왜냐하면 A는 2차 천이 과정에서 관목림을 거쳐 더 발전하면 양수림이 될 테니까. 따라서 ㉠은 양수림, ㉡은 지의류가 돼. 그러므로 보기의 ㄴ은 '㉠은 양수림이다'라고 했으니 맞는 설명이야.

그 다음으로 (나) 그래프를 보자. 식물 군집 K의 시간(t)에 따른 총생산량과 호흡량의 변화인데, t0 시점에서 출현하고, t1까지는 총생산량이 증가하다가 이후 감소하며, 호흡량은 계속 증가하고, t2 시점부터는 음수림 출현이라고 표시되어 있어. 즉, t1 이전에는 총생산량이 호흡량보다 많아서 순생산량이 양의 값을 가지다가, t1 이후에는 총생산량이 감소하면서 호흡량과 비슷해지거나 오히려 작아져서 순생산량이 줄어들게 되지. 

순생산량은 총생산량 - 호흡량이니까, t1 때는 총생산량이 최대이고 호흡량은 상대적으로 낮아서 순생산량이 매우 높을 거야. 그러나 t2 때가 되면 총생산량은 더 낮아지고 호흡량은 계속 증가해서 순생산량이 t1 때보다 훨씬 낮아질 거야. 따라서 K의 순생산물 / 호흡량 비율은 t1 때가 t2 때보다 크다는 결론이 나와. 그런데 문제의 보기 ㄷ은 'K의 순생산량/호흡량은 t2일 때가 t1일 때보다 크다'라고 했으므로 이는 틀린 설명이야. 따라서 ㄷ도 틀렸다.

정리하면:

- ㄱ: B는 2차 천이이다. → B는 1차 천이이므로 틀림.

- ㄴ: ㉠은 양수림이다. → 맞음.

- ㄷ: 순생산량/호흡량은 t2 > t1 → 틀림. t1이 더 큼.

따라서 정답은 ㄴ만 맞는 경우인데, 보기 중에 ㄴ만 있는 선택지가 있는지 확인해봐야 해. 문제에서 제시된 보기 옵션은 다음과 같아:

① ㄱ  

② ㄴ  

③ ㄱ, ㄷ  

④ ㄴ, ㄷ  

⑤ ㄱ, ㄴ, ㄷ  

여기서 ㄴ만 맞는 경우는 ②번이지만, 혹시 계산 실수가 있었는지 다시 확인해보자.

t1 시기에는 총생산량이 높고 호흡량은 아직 낮은 상태라 순생산량이 많을 거고, t2 시기에는 총생산량이 감소했고 호흡량은 증가했으므로 순생산량은 더 적어져. 따라서 순생산량 대비 호흡량의 비율, 즉 순생산량/호흡량은 t1 때가 훨씬 높다. 예를 들어 t1에서 총생산량이 10, 호흡량 3이면 순생산량은 7이고 비율은 7/3 ≈ 2.33. t2에서 총생산량 5, 호흡량 4라면 순생산량 1, 비율은 1/4 = 0.25. 따라서 확실히 t1이 더 크다. 그러니 ㄷ은 틀렸고, 따라서 정답은 ②번.

**Final Answer**:
\boxed{(2)}
\texttt{<|im\_end|>}

\texttt{<|im\_start|>assistant}

그림 (가)와 (나)를 분석하여 문제를 해결한다.

1. **천이 과정 분석**:

   - **A**: 초원에서 시작하여 관목림을 거쳐 ㉠으로 진행된다. 이미 식물이 있는 초원에서 시작되므로 **2차 천이**이다.
   
   - **B**: 용암 대지에서 시작하여 ㉡을 거쳐 초원으로 진행된다. 토양이 없는 용암 대지에서 시작하므로 **1차 천이**이다.
   
   - 따라서 ㄱ("B는 2차 천이이다.")은 **틀린** 설명이다.

2. **㉠과 ㉡ 식별**:

   - **A**(2차 천이): 관목림을 거쳐 ㉠으로 진행되며, 일반적으로 2차 천이의 후기 단계는 **양수림**이다.
   
   - **B**(1차 천이): 용암 대지에서 시작하여 ㉡을 거쳐 초원으로 진행되며, 초기 단계는 **지의류**이다.
   
   - 따라서 ㄴ("㉠은 양수림이다.")은 **맞는** 설명이다.

3. **그래프 분석**:

   - **총생산량**은 \( t_1 \)까지 증가한 후 감소하며, **호흡량**은 지속적으로 증가한다.
   
   - **순생산량**은 \( t_1 \) 시기에 총생산량이 최대이고 호흡량이 상대적으로 낮아 비율이 높다. \( t_2 \) 시기에는 총생산량이 감소하고 호흡량이 증가하여 순생산량 비율이 낮아진다.
   
   - 따라서 ㄷ("\( K \)의 \( \frac{\text{순생산량}}{\text{호흡량}} \)은 \( t_2 \)일 때가 \( t_1 \)일 때보다 크다")은 **틀린** 설명이다.

최종적으로 옳은 설명은 **ㄴ**뿐이다.

\boxed{(2)}
\texttt{<|im\_end|>}
\end{tcolorbox}

\subsection{KCSAT Biology Advanced Example}

\begin{figure}[ht!]
    \centering
    \includegraphics[width=0.65\linewidth]{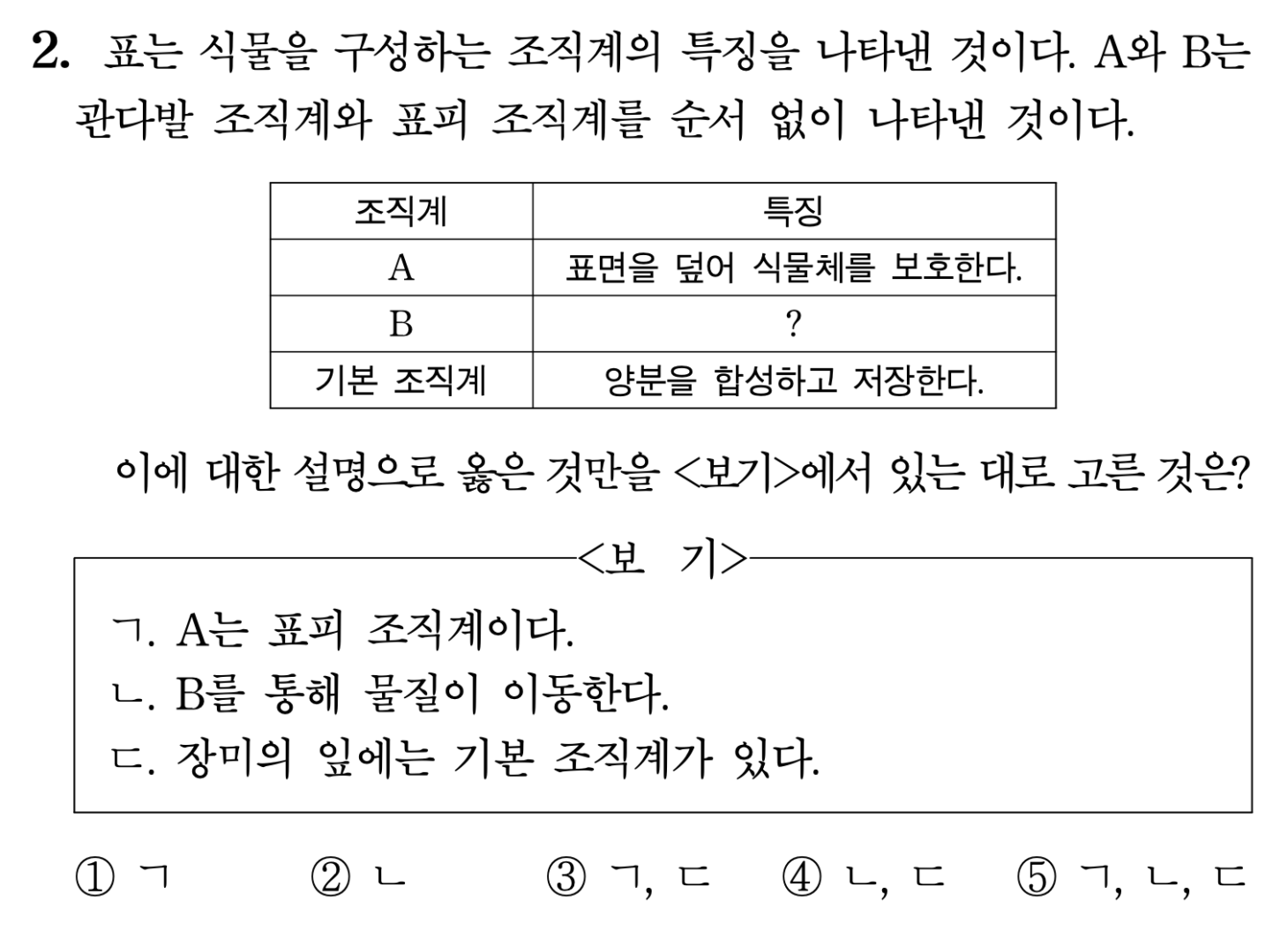}
    \caption{Original KCSAT Sample (in Korean).}
    \label{fig:kcsat-biology-png}
\end{figure}

We present our model’s output on a sample KCSAT Biology Advanced problem. Figure~\ref{fig:kcsat-biology-png} shows the original problem image, which is directly provided to the model for analysis and reasoning. The model is required to interpret, reason through, and solve the problem solely based on the image. For reference, we include the model’s original answer, as well as an English translation of the problem for readers. A literal English translation of the model’s output is also provided for comparison.


\begin{tcolorbox}[colback=gray!3!white, colframe=gray!80!black, boxrule=0.3mm, width=0.97\linewidth]
\textbf{Sample Question 2 (Translated)}

The table below summarizes the characteristics of plant tissues. A and B refer to vascular and dermal tissues, listed in no particular order.

\begin{center}
\begin{tabular}{|c|l|}
\hline
\textbf{Tissue}      & \textbf{Characteristic}                         \\ \hline
A                    & Covers the surface and protects the plant body. \\ \hline
B                    & ?                                               \\ \hline
Fundamental tissue   & Synthesizes and stores nutrients.               \\ \hline
\end{tabular}
\end{center}

Which statements below, from the options labeled (a) to (c), are correct regarding the tissues described above?

\medskip

\noindent
\textbf{(Options)}

\medskip
\fbox{%
  \parbox{0.93\linewidth}{%
    (a) A is a dermal tissue.\\
    (b) Substances are transported via tissue B.\\
    (c) The petal of a rose contains fundamental tissue.%
  }%
}

\medskip

Select the correct combination(s):

\begin{center}
\begin{tabular}{cccccc}
(1) a         & (2) b         & (3) a, c    & (4) b, c    & (5) a, b, c \\
\end{tabular}
\end{center}

\end{tcolorbox}

\begin{tcolorbox}[colback=cyan!1!white, colframe=gray!10!cyan,
  breakable, 
  title=HyperCLOVA X THINK (Translated Model Output -- English)]
  
\texttt{<|im\_start|>assistant/think}

Okay, let’s solve this problem. First, I need to carefully read the question. The question gives a table showing the characteristics of plant tissue systems, and A and B are listed as vascular tissue system and epidermal tissue system in random order. Also, it is stated that the ground tissue system synthesizes and stores nutrients.

First, looking at the table, the characteristic of A is “covers the surface and protects the plant body,” and B is not yet explained. However, since the ground tissue system is already described as synthesizing and storing nutrients, the remaining two must be the vascular and epidermal tissue systems. The problem states that one of A or B is epidermal, and the other is vascular.

Let’s summarize the characteristics of each tissue system.

1. Ground tissue system: This is also known as the fundamental tissue and mainly consists of parenchyma and sclerenchyma, which do not perform photosynthesis but serve to store or transport water and nutrients. So, in the problem, the characteristic of the ground tissue system is given as “synthesizes and stores nutrients”… Wait, does the ground tissue system photosynthesize? No, usually it’s tissues like palisade cells in the epidermal tissue or vascular tissue that do that, isn’t it?
Wait, this could be a confusing point. I was taught that the ground tissue system generally does not photosynthesize, but in the problem, its characteristic is “synthesizes and stores nutrients.” I need to check if this is correct.

Is the problem wrong? Or am I misremembering?

The ground tissue system mainly stores, and photosynthesis occurs in the epidermal tissue (especially palisade cells below the leaf epidermis). So is the description of the ground tissue system in the problem incorrect? Or is there another interpretation?

Since the problem states that the ground tissue system “synthesizes and stores nutrients,” I have no choice but to accept it as it is for now. Probably, that’s how the problem defines it.

Going back to the question: is A the epidermal tissue system or vascular? The characteristic of A is “covers the surface and protects the plant body,” which is a major function of the epidermal tissue system. For example, epidermal cells cover the surface of leaves, preventing water evaporation and protecting from pathogens. Therefore, A is epidermal tissue system, and B is likely vascular tissue system.

Then, what would be the feature of B? The vascular tissue system is composed of xylem and phloem and is involved in the transport of water and nutrients, so the characteristic of B would be something like “transports water and nutrients.” But the characteristic of B is left blank in the problem, so we have to infer it and select the correct explanation from the choices.

Now let’s look at the choices.

ㄱ. A is the epidermal tissue system. → Correct. Based on the reasoning above, this is true.

ㄴ. Substances move through B. → If B is the vascular tissue system, then it is the pathway for water and nutrient transport, so this is also true.

ㄷ. Rose leaves have ground tissue system. → The ground tissue system should exist in all plant leaves. As the internal tissue of leaves, it stores or transports water and nutrients, so this is also true.

However, the problem describes the ground tissue system as “synthesizes and stores nutrients,” which makes me doubtful as to whether it really does photosynthesis, but since that is what the problem states, I have to accept it. So ㄷ is also a correct explanation.

But based on prior knowledge, the ground tissue system does not photosynthesize, so I wonder if the problem is in error. If, for example, the problem actually refers to the epidermal tissue system instead? No, the problem specifically describes the characteristic of the ground tissue system, so I should just follow that.

So are ㄱ, ㄴ, ㄷ all correct? The answer options include 5 as “ㄱ, ㄴ, ㄷ,” so that seems possible, but is there any chance I made a mistake?

Let me check again.

A: Epidermal tissue system → covers surface. O

B: Vascular tissue system → transports water and nutrients. O

Ground tissue system: In problem, synthesizes and stores nutrients. Might be different from reality but will follow the problem.

Do rose leaves have ground tissue system? The inside of leaves is made of parenchyma, which is ground tissue, so O.

Therefore, it seems all three are correct. Is the answer 5?

But sometimes in tests there are traps—could ㄷ somehow be wrong?

For example, if rose leaves had only vascular tissues and not ground tissue? But in a leaf, there are epidermis, guard cells, palisade tissue (ground tissue), vascular bundles, etc., so ground tissue must be present. Thus, ㄷ is valid.

So the answer is 5?

Even though at first I thought the description of the ground tissue system was wrong, since that’s how the problem gives it, I will follow it. Therefore, all ㄱ, ㄴ, and ㄷ are correct, and the answer is 5.

**Final answer**:
\boxed{(5)}
\texttt{<|im\_end|>}

\texttt{<|im\_start|>assistant}

In the table for question 2, A and B represent the epidermal tissue system and vascular tissue system, respectively. The characteristic of the ground tissue system is given as 'synthesizes and stores nutrients.'

- **Characteristic of A**: 'Covers the surface and protects the plant body.'  
  This is a main function of the epidermal tissue system, which covers the surface of the plant to prevent water evaporation and protect from external environments. Therefore, A is the epidermal tissue system.

- **Characteristic of B**: It is a characteristic of the vascular tissue system. The vascular tissue system is composed of xylem and phloem and is involved in the transport of water and nutrients. Therefore, B is the vascular tissue system, and it is correct to say that substances are transported through it.

- **Ground tissue system**: The characteristic given in the problem is 'synthesizes and stores nutrients.'  
  The ground tissue system comprises most of the plant and serves to store and transport water and nutrients. The leaves of a rose also contain ground tissue system, carrying out these functions.

Analysis of choices:

- **ㄱ**: A is the epidermal tissue system. (Correct)

- **ㄴ**: Substances are transported through B. (Correct)

- **ㄷ**: The leaves of a rose have ground tissue system. (Correct)

Therefore, since all choices are correct, the answer is number 5.

\boxed{(5)}
\texttt{<|im\_end|>}
\end{tcolorbox}

\begin{tcolorbox}[colback=cyan!1!white, colframe=gray!10!cyan,
  breakable, 
  title=HyperCLOVA X THINK (Original Model Output -- Korean)]
\texttt{<|im\_start|>assistant/think}

좋아, 이 문제를 풀어보자. 우선 문제를 잘 읽어봐야겠어. 문제는 식물을 구성하는 조직계의 특징을 나타낸 표인데, A와 B는 관다발 조직계와 표피 조직계를 순서 없이 나열했다고 해. 그리고 기본 조직계는 양분을 합성하고 저장한다고 되어 있네.

먼저, 표를 보면 A의 특징은 '표면을 덮어 식물체를 보호한다'고 했고, B는 아직 설명이 안 되어 있어. 그런데 기본 조직계는 이미 양분 합성과 저장이라고 했으니까, 나머지 두 개가 관다발과 표피일 거야. 문제에서 A와 B 중 하나는 표피이고 다른 하나는 관다발이야.

그럼 먼저 각 조직계의 특징을 정리해보자. 

1. **기본 조직계**: 이건 울타리라고도 불리는 조직으로, 주로 유조직과 기계 조직으로 이루어져 있고, 광합성을 하지 않지만 물과 영양분을 저장하거나 운반하는 역할을 한다고 알고 있어. 그래서 문제에서도 기본 조직계의 특징이 '양분을 합성하고 저장한다'고 했는데... 잠깐, 기본 조직계가 광합성을 하나? 아니, 보통 표피 조직이나 관다발 조직 중에 책상세포 같은 게 있지 않나? 

잠깐, 여기서 헷갈리기 쉬운 부분이 있을 것 같아. 기본 조직계는 일반적으로 광합성을 하지 않는다고 배웠던 것 같은데, 문제에서는 기본 조직계의 특징이 '양분을 합성하고 저장한다'고 했어. 이게 맞는지 확인해야 할 것 같아. 

아니면 문제가 잘못된 건가? 아니면 내가 잘못 기억한 건가?

기본 조직계는 주로 저장 기능을 하며, 광합성은 표피 조직(특히 잎의 표피 아래에 있는 책상세포)에서 일어나는 걸로 알고 있어. 그럼 문제의 기본 조직계 설명이 틀린 건가? 아니면 다른 해석이 있을까?

문제에서 기본 조직계의 특징이 '양분을 합성하고 저장한다'고 했으니 일단 그대로 받아들이고 넘어가야겠지. 아마 문제에서는 그렇게 정의했으니까.

그럼 다시 문제로 돌아가서, A는 표피 조직계일까 관다발일까? A의 특징은 '표면을 덮어 식물체를 보호한다'인데, 표피 조직계의 주요 기능 중 하나가 바로 식물체의 표면 덮개 역할이지. 예를 들어 잎의 표피 세포들이 표면을 덮어서 수분 증발을 막고 병원체로부터 보호해주니까. 따라서 A는 표피 조직계이고, B는 관다발 조직계일 가능성이 높아.

그러면 B의 특징은 뭘까? 관다발 조직계는 물관부와 체관부로 나뉘어져 있어서 물과 양분의 이동에 관여하니까, B의 특징은 아마도 '물과 양분을 운반한다' 정도가 될 것 같아. 하지만 문제에서는 B의 특징이 비어있고, 우리가 그걸 추론해서 보기 중에서 맞는 설명을 골라야 하니까.

이제 보기를 살펴보자.

ㄱ. A는 표피 조직계이다. → 맞아. 위에서 생각한 대로라면 맞는 말이야.

ㄴ. B를 통해 물질이 이동한다. → B가 관다발 조직계라면, 물과 양분이 이동하는 통로이므로 맞는 말이지.

ㄷ. 장미의 잎에는 기본 조직계가 있다. → 기본 조직계는 모든 식물의 잎에 존재할 거야. 잎의 내부 조직으로서 물과 양분을 저장하거나 운반하는 역할을 하니까. 따라서 이것도 맞는 말이야.

그런데 문제에서 기본 조직계의 특징이 '양분을 합성하고 저장한다'고 했는데, 실제로 기본 조직계가 광합성을 하는지 의문이 들지만, 문제에서는 그렇게 주어졌으므로 일단은 받아들여야 해. 그러면 ㄷ도 맞는 설명이 되겠지.

하지만 기존 지식으로는 기본 조직계가 광합성을 하지 않는다고 알고 있는데, 이 부분이 문제의 오류인지 아닌지가 궁금해. 만약 문제에서의 기본 조직계가 실제로는 표피 조직계를 가리키는 것이 아니라면? 아니, 문제에서는 기본 조직계의 특징을 따로 주었으니 그대로 따라가야 할 것 같아.

따라서 ㄱ, ㄴ, ㄷ 모두 옳은 설명인가? 그런데 보기에서 5번이 'ㄱ, ㄴ, ㄷ'이라서 그럴 수도 있지만, 혹시 내가 뭔가 잘못 생각했을까?

다시 한 번 점검해보자.

- A: 표피 조직계 → 표면 덮개 역할. O

- B: 관다발 조직계 → 물과 양분 운반. O

- 기본 조직계: 문제에서는 양분 합성 및 저장. 실제와 다를 수 있으나 문제에 따름.

장미의 잎에는 기본 조직계가 있는가? 잎의 내부는 유조직으로 구성되어 있는데, 이는 기본 조직계에 해당하므로 O.

그래서 세 가지 모두 맞는 것 같다. 그렇다면 정답은 5번?

하지만 보통 시험 문제에서는 이런 함정이 있을 수 있으니, 혹시 ㄷ가 틀린 경우는 없을까?

예를 들어, 장미의 잎에는 기본 조직계가 없고 관다발만 있다는 식으로? 아니, 잎의 경우 표피, 공변세포, 울타리 조직(기본 조직계), 관다발 등이 있으니까 기본 조직계가 있어야 해. 따라서 ㄷ는 맞는 설명.

그럼 답은 5번?

하지만 처음에 기본 조직계의 특징이 틀렸다는 생각이 들었지만, 문제에서는 그렇게 주어졌으니 따라가야 한다. 따라서 ㄱ, ㄴ, ㄷ 모두 맞는 설명이므로 5번이 정답이다.

**최종 답변**
\boxed{(5)}
\texttt{<|im\_end|>}

\texttt{<|im\_start|>assistant}

2번 문제의 표에서 A와 B는 각각 표피 조직계와 관다발 조직계를 나타낸다. 기본 조직계의 특징은 '양분을 합성하고 저장한다'로 주어져 있다.

- **A의 특징**: '표면을 덮어 식물체를 보호한다.'  
  이는 표피 조직계의 주요 기능으로, 표피 조직계는 식물체의 표면을 덮어 수분 증발을 막고 외부 환경으로부터 보호한다. 따라서 A는 표피 조직계이다.

- **B의 특징**: 관다발 조직계의 특징이다. 관다발 조직계는 물관과 체관으로 구성되어 물과 양분의 이동에 관여한다. 따라서 B는 관다발 조직계이며, 이를 통해 물질이 이동한다는 것은 올바른 설명이다.

- **기본 조직계**: 문제에서 주어진 특징은 '양분을 합성하고 저장한다.'  
  기본 조직계는 식물의 대부분을 차지하며, 물과 양분의 저장 및 운반 역할을 한다. 장미의 잎에도 기본 조직계가 존재하여 이러한 기능을 수행한다.

보기 분석:
- **ㄱ**: A는 표피 조직계이다. (정답)
- **ㄴ**: B를 통해 물질이 이동한다. (정답)
- **ㄷ**: 장미의 잎에는 기본 조직계가 있다. (정답)

따라서 모든 선택지가 올바르므로 정답은 5번이다.

\boxed{(5)}
\texttt{<|im\_end|>}
\end{tcolorbox}

\subsection{KCSAT Math Advanced Example}

\begin{figure}[ht!]
    \centering
    \includegraphics[width=0.7\linewidth]{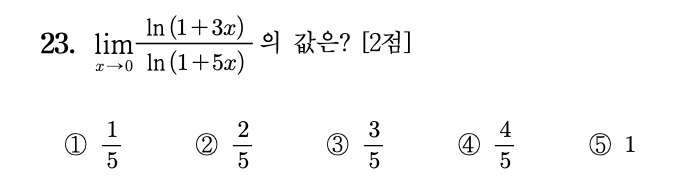}
    \caption{Original KCSAT Sample (in Korean).}
    \label{fig:kcsat-math-png}
\end{figure}

We present our model’s output on a sample KCSAT Math Advanced problem. Figure~\ref{fig:kcsat-math-png} shows the original problem image, which is directly provided to the model for analysis and reasoning. The model is required to interpret, reason through, and solve the problem solely based on the image. For reference, we include the model’s original answer, as well as an English translation of the problem for readers. A literal English translation of the model’s output is also provided for comparison.


\begin{tcolorbox}[colback=gray!3!white, colframe=gray!80!black, boxrule=0.3mm, width=0.95\linewidth, enlarge left by=0mm]
\textbf{Sample Question 23 (Translated)}

Determine the value of
\[
    \lim_{x \to 0} \frac{\ln(1+3x)}{\ln(1+5x)}
\]

\noindent
Choose the correct answer:  
(1) $\frac{1}{5}$ \qquad (2) $\frac{2}{5}$ \qquad (3) $\frac{3}{5}$ \qquad (4) $\frac{4}{5}$ \qquad (5) $1$
\end{tcolorbox}

\begin{tcolorbox}[colback=cyan!1!white, colframe=gray!10!cyan, 
  breakable, 
  title=HyperCLOVA X THINK (Translated Model Output -- English)]
\texttt{<|im\_start|>assistant/think}

Alright, this seems to be a problem about finding a limit. It’s question 23, and it presents the ratio of logarithmic functions in fractional form. The expression is the limit as $x \to 0$ of $\frac{\ln(1+3x)}{\ln(1+5x)}$. The provided choices range from $\frac{1}{5}$ to $1$, but which one is correct?

First, for limits of this type, you can usually use L'Hôpital’s Rule to solve them, since as $x \to 0$, both the numerator and denominator become $\ln(1)$, so the expression is in the indeterminate form $0/0$. By applying L'Hôpital’s Rule, we can find the answer by differentiating the numerator and denominator.

Before applying L'Hôpital’s Rule, let’s check: when $x=0$, the numerator is $\ln(1 + 3 \cdot 0) = \ln(1) = 0$, and the denominator is $\ln(1 + 5 \cdot 0)=\ln(1)=0$, so that checks out. Therefore, L'Hôpital’s Rule can be used.
So, we need to differentiate both the numerator and denominator with respect to $x$. The derivative of the numerator $\ln(1+3x)$ is $\frac{3}{1+3x}$, and the derivative of the denominator $\ln(1+5x)$ is $\frac{5}{1+5x}$.

By L'Hôpital’s Rule, the limit then becomes
\[
\frac{\frac{3}{1+3x}}{\frac{5}{1+5x}} = \frac{3}{1+3x} \cdot \frac{1+5x}{5}
\]
and, simplifying this,
\[
\frac{3(1+5x)}{5(1+3x)}
\]
is what you get.

Now, if you calculate the value as $x \to 0$, $1+5x \to 1$, and $1+3x \to 1$, so overall it should become $\frac{3}{5}$, right?
Looking at the choices, $\frac{3}{5}$ is choice ③. But why did I think of something else at first? Oh, did I calculate something incorrectly?

Let me double-check. After applying L'Hôpital’s Rule:

\[
\lim_{x \to 0} \frac{3}{1+3x} \cdot \frac{1+5x}{5}
\]

Plugging in $x \to 0$,

\[
\frac{3}{1} \cdot \frac{1}{5} = \frac{3}{5}
\]

So, the answer is $\frac{3}{5}$? Among the answer choices, ③ is $\frac{3}{5}$, ② is $\frac{2}{5}$, and ④ is $\frac{4}{5}$, so I need to make sure not to get confused.

Can this be solved another way, for example, by approximating using a Taylor expansion?

\[
\ln(1+a) \approx a - \frac{a^2}{2} + \cdots
\]

Using this expansion, if $a$ is small, we can approximate $\ln(1+a) \approx a$.

So, in the numerator, $\ln(1+3x) \approx 3x$, and in the denominator, $\ln(1+5x) \approx 5x$, which makes the ratio
\[
\frac{3x}{5x} = \frac{3}{5}
\]
and we reach the same result.

So, since both methods yield $\frac{3}{5}$, it seems choice ③ is the answer.
Is there any chance I made a calculation mistake in the process? Let’s check again.
When applying L'Hôpital’s Rule, did I correctly differentiate the numerator and denominator:

\[
\frac{d}{dx} \ln(1+3x) = \frac{3}{1+3x}, \quad \frac{d}{dx} \ln(1+5x) = \frac{5}{1+5x}
\]

And when converting the division to multiplication, there was no sign or coefficient error, and as $x \to 0$, the $x$ terms in the denominator and numerator disappear, so the result is indeed $\frac{3}{5}$.
In the case of the Taylor expansion as well, considering only the first term, it’s exactly $\frac{3x}{5x} = \frac{3}{5}$. So, certainly, $\frac{3}{5}$ is correct.
Therefore, among the choices, since ③ is $\frac{3}{5}$, the final answer is ③.

**Final answer**:
\boxed{(3)}
\texttt{<|im\_end|>}

\texttt{<|im\_start|>assistant}

To find the limit $\lim_{x \to 0} \frac{\ln(1+3x)}{\ln(1+5x)}$, we use the **L’Hôpital’s Rule**.

Since both the numerator and denominator become the undetermined form $0/0$ as $x \to 0$, we can apply L’Hôpital’s Rule:
\[
\lim_{x \to 0} \frac{\ln(1+3x)}{\ln(1+5x)} = \lim_{x \to 0} \frac{\frac{d}{dx} \ln(1+3x)}{\frac{d}{dx} \ln(1+5x)}
\]

Differentiating each:
\[
\frac{d}{dx} \ln(1+3x) = \frac{3}{1+3x}, \quad \frac{d}{dx} \ln(1+5x) = \frac{5}{1+5x}
\]

Therefore,
\[
\lim_{x \to 0} \frac{\frac{3}{1+3x}}{\frac{5}{1+5x}} = \lim_{x \to 0} \frac{3}{1+3x} \cdot \frac{1+5x}{5}
\]

When $x \to 0$, $1+3x \approx 1$ and $1+5x \approx 1$, so
\[
\frac{3}{1} \cdot \frac{1}{5} = \frac{3}{5}
\]

Also, using a Taylor expansion, $\ln(1 + ax) \approx ax$ (where $a$ is a constant) as an approximation:
\[
\frac{\ln(1+3x)}{\ln(1+5x)} \approx \frac{3x}{5x} = \frac{3}{5}
\]

Therefore, the answer is $\boxed{(3)}$.

\texttt{<|im\_end|>}
\end{tcolorbox}

\begin{tcolorbox}[colback=cyan!1!white, colframe=gray!10!cyan, 
  breakable, 
  title=HyperCLOVA X THINK (Original Model Output -- Korean)]

\texttt{<|im\_start|>assistant/think}

좋아, 이 문제는 극한을 구하는 문제인 것 같아. 23번 문제고, 분수꼴의 로그 함수의 비율이 주어졌어. 식은 $x \to 0$일 때 $\frac{\ln(1+3x)}{\ln(1+5x)}$의 극한이야. 보기로는 $\frac{1}{5}$부터 $1$까지 있는데, 정답이 뭘까?

먼저, 이런 형태의 극한 문제는 보통 로피탈 법칙을 써서 풀 수 있지 않을까? 왜냐하면 분자와 분모 모두 $x \to 0$일 때 $\ln(1)$이 되니까 $0/0$ 형태의 불정형이기 때문이야. 로피탈 법칙을 적용하면 미분해서 계산할 수 있을 거야.

로피탈 법칙을 적용하기 전에 먼저 확인해보자. $x=0$일 때 분자는 $\ln(1 + 3 \cdot 0) = \ln(1) = 0$, 분모도 $\ln(1 + 5 \cdot 0)=\ln(1)=0$이니까 맞아. 따라서 로피탈 법칙을 사용할 수 있어.
그래서 분자와 분모를 각각 $x$로 미분해야 해. 
분자인 $\ln(1+3x)$의 도함수는 $\frac{3}{1+3x}$이고, 분모인 $\ln(1+5x)$의 도함수는 $\frac{5}{1+5x}$이지.

그러면 로피탈 법칙에 의해 극한은
\[
\frac{\frac{3}{1+3x}}{\frac{5}{1+5x}} = \frac{3}{1+3x} \cdot \frac{1+5x}{5}
\]
가 되고, 이걸 단순화하면
\[
\frac{3(1+5x)}{5(1+3x)}
\]
가 되겠지.

이제 $x \to 0$일 때 이 값을 계산하면, $1+5x \to 1$, $1+3x \to 1$이니까 전체적으로 $\frac{3}{5}$가 되어야 하지 않나?
근데 보기에는 $\frac{3}{5}$가 ③번에 있네. 그런데 왜 내가 처음 생각했을 때 다른 결과가 나오는 걸까? 아, 아니, 계산을 잘못했나?

다시 확인해보자. 로피탈 법칙 적용 후의 식:

\[
\lim_{x \to 0} \frac{3}{1+3x} \cdot \frac{1+5x}{5}
\]

여기서 $x \to 0$을 대입하면,

\[
\frac{3}{1} \cdot \frac{1}{5} = \frac{3}{5}
\]

맞아, 그럼 답은 $\frac{3}{5}$인가? 그런데 문제에 주어진 보기 중에서 ③번이 $\frac{3}{5}$이고, ②번이 $\frac{2}{5}$, ④번이 $\frac{4}{5}$라서 헷갈리지 않게 해야겠어.

혹시 다른 방법으로 풀어보면, 예를 들어 테일러 전개를 사용해서 근사할 수 있을까?

\[
\ln(1+a) \approx a - \frac{a^2}{2} + \cdots
\]

이런 식으로 전개하면, $a$가 작을 때 $\ln(1+a) \approx a$ 정도로 근사할 수 있지.

그럼 분자는 $\ln(1+3x) \approx 3x$, 분모는 $\ln(1+5x) \approx 5x$가 되니까, 그 비율은
\[
\frac{3x}{5x} = \frac{3}{5}
\]
가 되어 동일한 결과를 얻을 수 있겠네.

그래서 두 방법 모두 $\frac{3}{5}$가 나오니까 정답은 ③번인 것 같다.
하지만 혹시나 계산 과정에서 실수한 부분이 있을까? 다시 한번 체크해보자.
로피탈 법칙 적용 시, 분자와 분모의 도함수를 제대로 했는지:

\[
\frac{d}{dx} \ln(1+3x) = \frac{3}{1+3x}, \quad \frac{d}{dx} \ln(1+5x) = \frac{5}{1+5x}
\]

그리고 나눗셈을 곱셈으로 바꿀 때 부호나 계수가 틀린 건 없고, $x \to 0$일 때의 값은 분모와 분자의 $x$ 항이 없어지므로 $\frac{3}{5}$가 되는 게 맞지.
테일러 전개의 경우에도 첫 번째 항만 보면 정확히 $\frac{3x}{5x} = \frac{3}{5}$. 그래서 확실히 $\frac{3}{5}$가 맞는 것 같아.
그럼 문제의 보기 중에서 ③번이 $\frac{3}{5}$이므로 정답은 ③번이다.

**최종 답변**
\boxed{(3)}\texttt{<|im\_end|>}

\texttt{<|im\_start|>assistant}

극한 $\lim_{x \to 0} \frac{\ln(1+3x)}{\ln(1+5x)}$ 를 구하기 위해 **로피탈 법칙**을 사용합니다.

분자와 분모 모두 $x \to 0$ 일 때 $0/0$ 형태의 부정형이므로 로피탈 법칙을 적용합니다:
\[
\lim_{x \to 0} \frac{\ln(1+3x)}{\ln(1+5x)} = \lim_{x \to 0} \frac{\frac{d}{dx} \ln(1+3x)}{\frac{d}{dx} \ln(1+5x)}
\]

각각 미분하면:
\[
\frac{d}{dx} \ln(1+3x) = \frac{3}{1+3x}, \quad \frac{d}{dx} \ln(1+5x) = \frac{5}{1+5x}
\]

따라서,
\[
\lim_{x \to 0} \frac{\frac{3}{1+3x}}{\frac{5}{1+5x}} = \lim_{x \to 0} \frac{3}{1+3x} \cdot \frac{1+5x}{5}
\]

$x \to 0$ 일 때 $1+3x \approx 1$ 및 $1+5x \approx 1$ 이므로,
\[
\frac{3}{1} \cdot \frac{1}{5} = \frac{3}{5}
\]

또한 테일러 전개를 사용하면 $\ln(1 + ax) \approx ax$ (단, $a$ 는 상수)로 근사할 수 있습니다:
\[
\frac{\ln(1+3x)}{\ln(1+5x)} \approx \frac{3x}{5x} = \frac{3}{5}
\]

따라서, 답은 $\boxed{(3)}$ 입니다.

\texttt{<|im\_end|>}
\end{tcolorbox}

\clearpage
\newpage
\section{Synthetic Data Examples} \label{appendix:synthetic_example}

\begin{figure}[ht]
    \centering
    \includegraphics[width=1\linewidth]{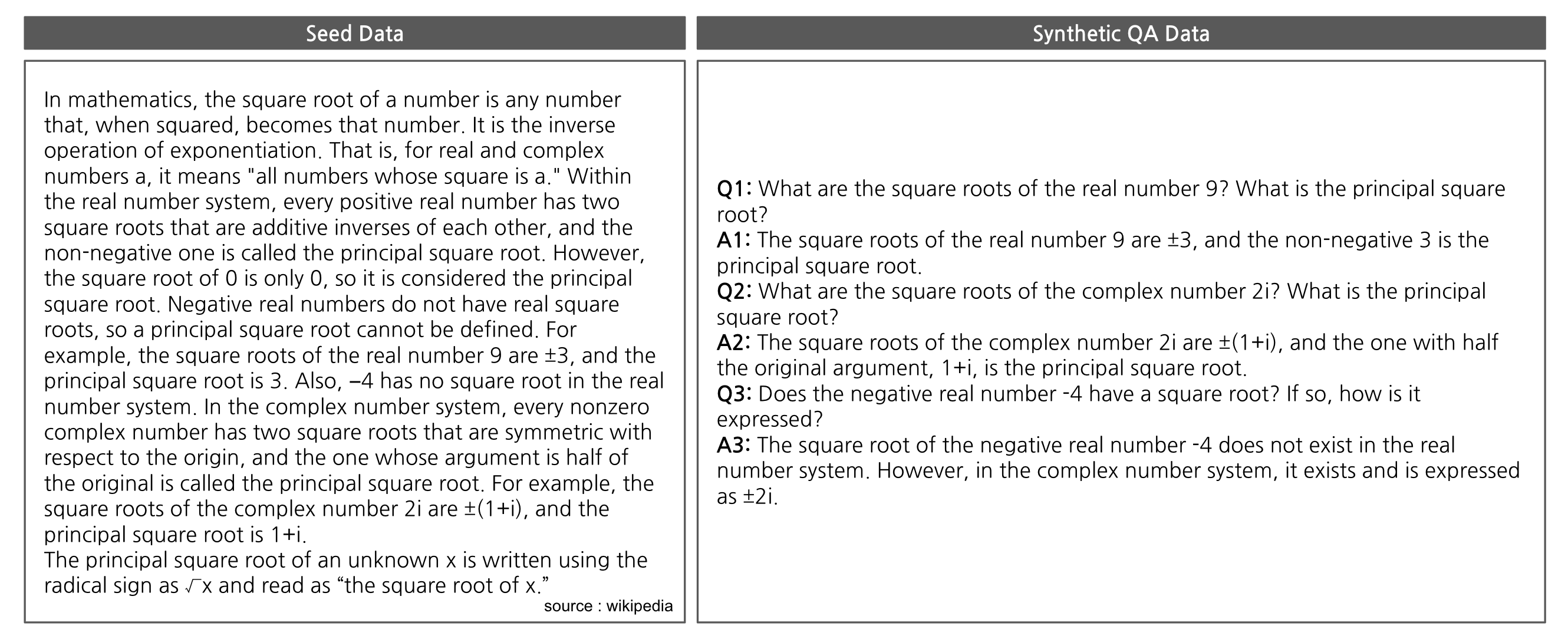}
    \caption{Synthetic QA data generated from Wikipedia}
\end{figure}
\begin{figure}[hp!]
    \centering
    \includegraphics[width=1\linewidth]{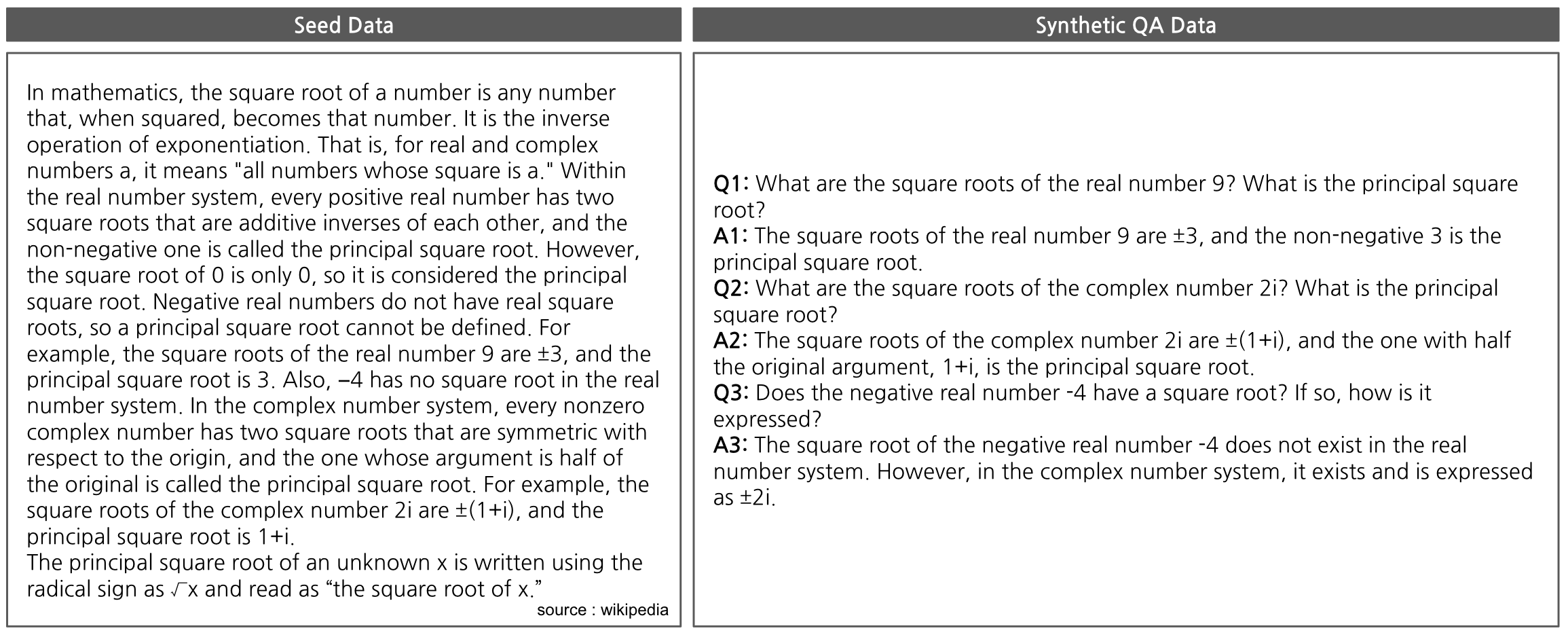}
    \caption{Translated sample example of synthetic QA data generated from Wikipedia}
\end{figure}
\begin{figure}[ht]
    \centering
    \includegraphics[width=1\linewidth]{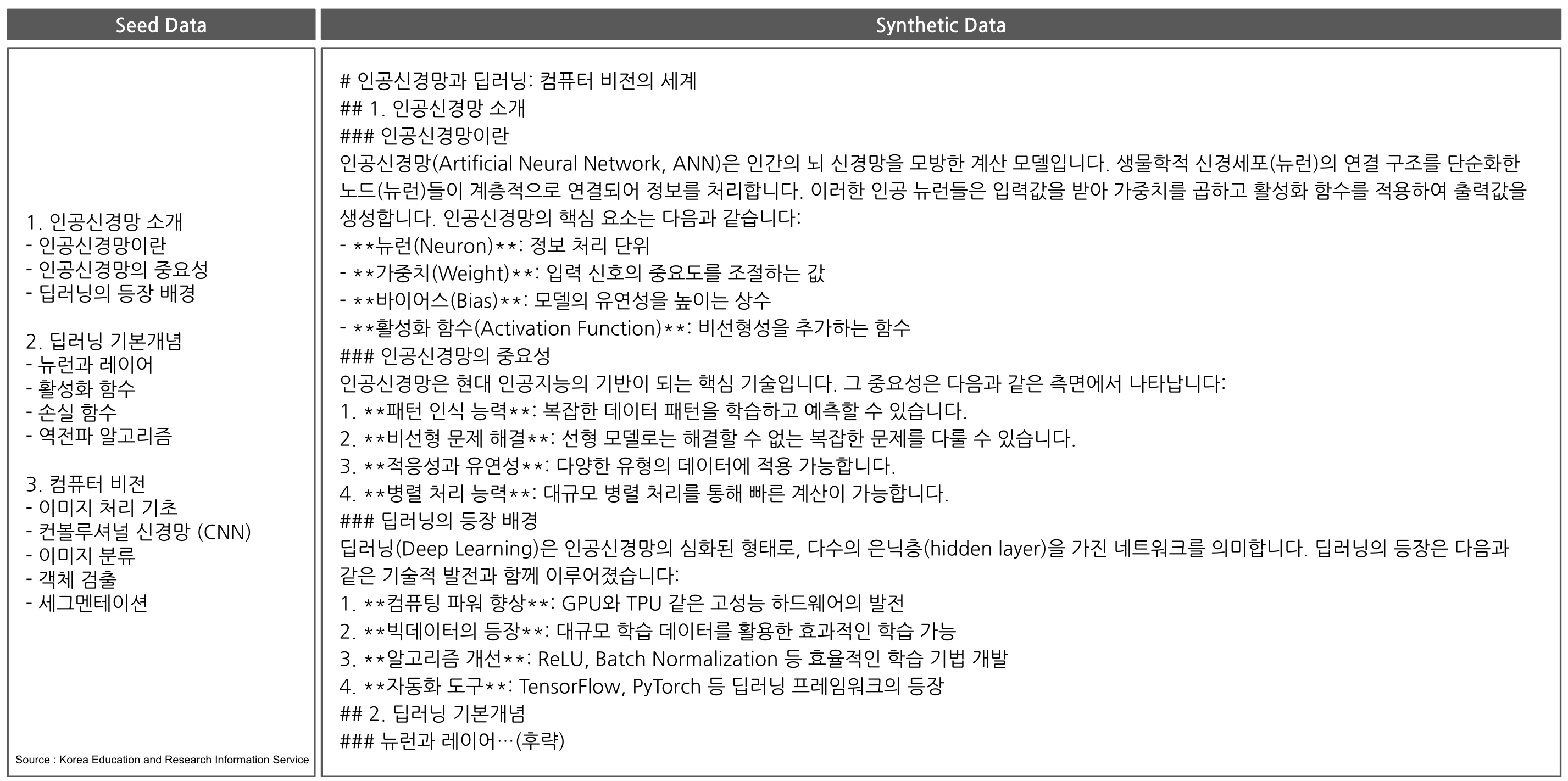}
    \caption{Synthetic educational data generated from course syllabus}
\end{figure}
\begin{figure}[ht]
    \centering
    \includegraphics[width=1\linewidth]{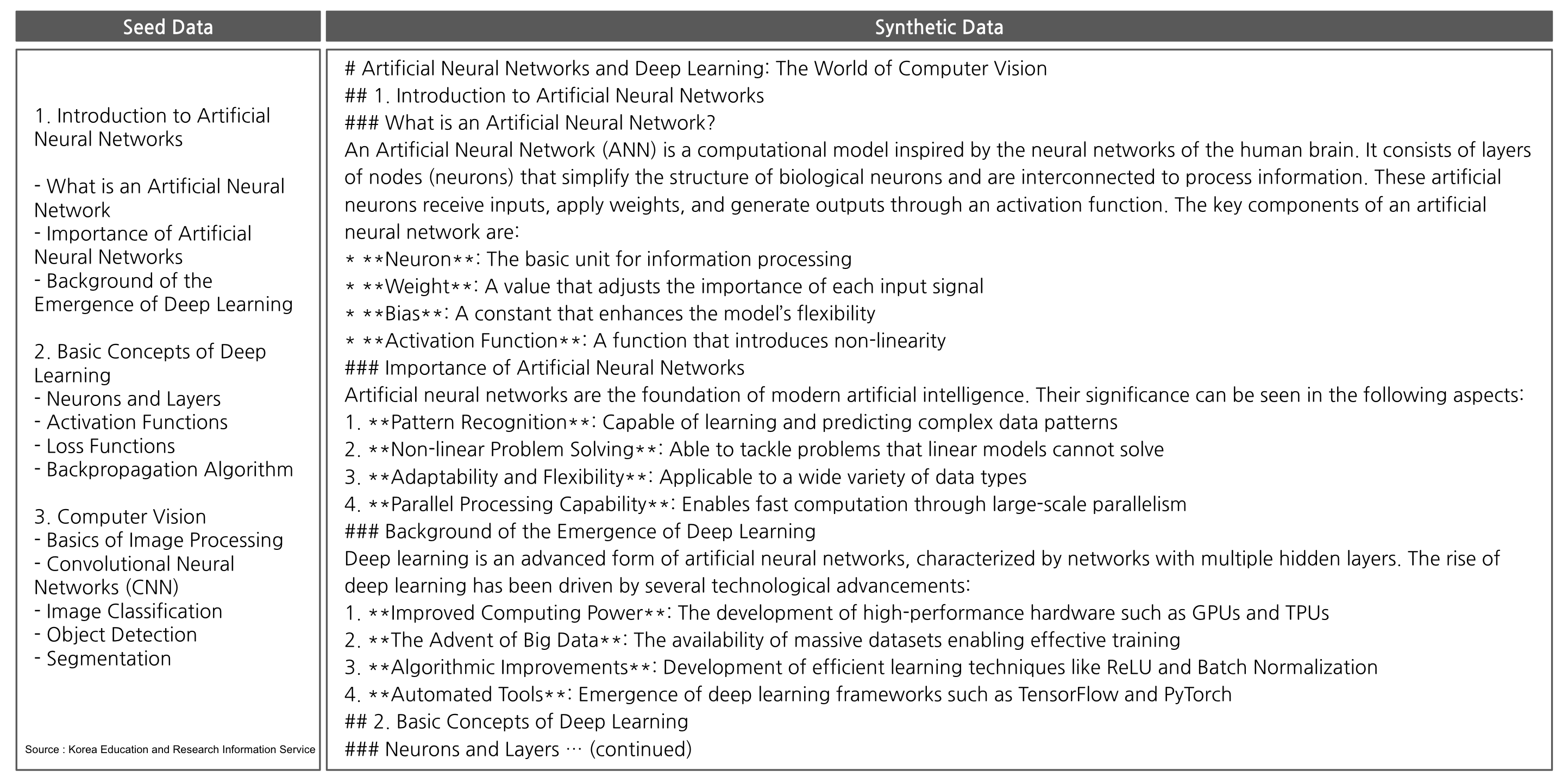}
    \caption{Translated sample example of synthetic educational data generated from course syllabus}
\end{figure}

\end{document}